\documentclass[]{visionxlab}

\usepackage[toc,page,header]{appendix}

\usepackage{makecell}
\usepackage{times}
\usepackage{latexsym}
\usepackage{enumitem}

\usepackage[T1]{fontenc}
\usepackage{hyperref} 
\usepackage{url}

\usepackage[utf8]{inputenc}

\usepackage{microtype}

\usepackage{inconsolata}

\usepackage{graphicx}
\usepackage{CJK}
\usepackage[dvipsnames,table]{xcolor}
\usepackage{xspace}
\usepackage{booktabs}
\usepackage{amsfonts} 
\usepackage{amssymb}
\usepackage{multirow}
\usepackage{multicol}
\usepackage{pifont}
\usepackage{bbding}

\usepackage{listings}
\usepackage{xcolor}
\usepackage{upquote}
\usepackage{mdframed}

\lstdefinestyle{promptstyle}{
  basicstyle=\ttfamily\footnotesize,
  breaklines=true,
  breakautoindent=false,
  breakindent=0pt,
  breakatwhitespace=false,
  columns=fullflexible,
  keepspaces=true,
  showstringspaces=false,
  extendedchars=true,
  inputencoding=utf8,
  literate=%
    {—}{{\textemdash}}1
    {≤}{{$\leq$}}1
    {≥}{{$\geq$}}1
    {→}{{$\rightarrow$}}1
    {⇌}{{$\rightleftharpoons$}}1
    {↑}{{$\uparrow$}}1
    {↓}{{$\downarrow$}}1
    {×}{{$\times$}}1
    {∅}{{$\emptyset$}}1
    {Δ}{{$\Delta$}}1
    {…}{{\ldots}}1
    {’}{{'}}1
    {‘}{{'}}1
}

\mdfdefinestyle{promptframe}{
  backgroundcolor=black!4,
  linecolor=black!40,
  linewidth=0.5pt,
  innerleftmargin=6pt,
  innerrightmargin=6pt,
  innertopmargin=5pt,
  innerbottommargin=5pt,
  skipabove=\baselineskip,
  skipbelow=\baselineskip,
}

\lstnewenvironment{promptlisting}{\lstset{style=promptstyle}}{}
\surroundwithmdframed[style=promptframe]{promptlisting}

\newcommand{\datasetname}{DisciplineGen-1M\xspace}

\title{\datasetname: A Large-Scale Dataset \\ for Multidisciplinary Visual Generation \\ and Editing}

\author[1,*,\ddagger]{Zhaokai Wang}
\author[1,*]{Mingxin Liu}
\author[1,*]{Zirun Zhu}
\author[2,*]{Ziqian Fan}
\author[1,*]{Yiguo He}
\author[3]{Mohan Zhang}
\author[1]{Leyao Gu}
\author[1]{Yan Li}
\author[1]{Xiangyu Zhao}
\author[1]{Ning Liao}
\author[4]{Shaofeng Zhang}
\author[1]{Xuanhe Zhou}
\author[1]{Zhihang Zhong}
\author[1,\dagger]{Xue Yang}

\affiliation[1]{Shanghai Jiao Tong University}
\affiliation[2]{South China University of Technology}

\affiliation[3]{Xiamen University}
\affiliation[4]{University of Science and Technology of China}

\contribution[*]{Equal Contribution}
\contribution[\ddagger]{Project Lead}
\contribution[\dagger]{Corresponding Author}

\abstract{
Recent image generation and editing models can produce visually appealing natural images, yet they remain unreliable when the target image is a knowledge-intensive diagram whose correctness depends on disciplinary concepts, symbolic structure, and precise spatial relations. We introduce \datasetname, a million-scale multidisciplinary dataset that supports text-to-image generation and image editing. It contains 1.2M samples spanning mathematics, physics, chemistry, biology, geography, computer science, economics, history, music, and sports. To construct the dataset, we design a scalable framework that combines structured rendering, OCR-based editing, specialized programmatic synthesis, and large-scale text-to-image filtering. These pipelines produce captions, editing instructions, structured annotations, and paired images with controllable semantic differences. Building on \datasetname, we further introduce a discipline-informed reasoning-generation model for both text-to-image generation and image editing. Experiments on discipline-related benchmarks, GenExam and GRADE, show substantial improvements over open-source baselines, while evaluations on general reasoning-informed benchmarks, WISE and RISE, further indicate broader transfer. The results suggest that large-scale structured academic visual data is a key ingredient for moving image generation from aesthetic plausibility toward verifiable knowledge-grounded visual creation. We will publicly release our dataset, model, and source code of the data curation pipeline to ensure reproducibility and benefit future research.
}

\date{\today}
\checkdata[Project Page]{\url{https://disciplinegen.github.io/}}

\setsjtublue

\begin{document}
\maketitle

\section{Introduction}

Existing generative models have made remarkable progress in image generation and editing, demonstrating the capability to synthesize images with high aesthetic quality and strong alignment with text prompts. Concurrently, with the evolution of generative models, reasoning-informed generation has emerged as a prevailing trend and a highly active research direction. By unifying generation and understanding, this paradigm fosters mutual advancement alongside the development of unified multimodal models. Although several benchmarks have been established for reasoning-informed generation, they reveal that current generative models still exhibit limited performance in scenarios requiring precise comprehension, complex structural analysis, and deep reasoning. However, there remains a critical shortage of datasets dedicated to assisting and supporting this specific scenario.

\begin{table*}[t]
\small
\setlength{\tabcolsep}{3pt}
\resizebox{1.0\textwidth}{!}{
    \centering
    \renewcommand{\arraystretch}{1.5}

    \begin{tabular}{lccccccc}
    \toprule
    \multirow{1}{*}{Dataset} & \multirow{1}{*}{Domain} & \multirow{1}{*}{\#Samples} & \multirow{1}{*}{Format} & \multirow{1}{*}{Source} & \makecell[c]{Structured\\Image} &
    \makecell[c]{Multi-\\Discipline} &
    \makecell[c]{Reasoning /\\Knowledge}\\
    \midrule

    BizGen~\cite{peng2025bizgenadvancingarticlelevelvisual} & \makecell[c]{Structured + \\Text-rich} & 650K & T2I & \makecell[c]{Synthetic +\\Retrieval} & \Checkmark & \XSolidBrush & Partial \\
    VectorEdits~\cite{kuchar2025vectoreditsdatasetbenchmarkinstructionbased} & \makecell[c]{Structured} & 271K & Edit & Real & \Checkmark & \XSolidBrush & Partial \\
    StructVisuals~\cite{zhuo2025factuality} & \makecell[c]{Scientific / Structured} & 1.3M & \makecell[c]{T2I + Edit} & Programmatic & \Checkmark & \XSolidBrush & \Checkmark \\
    TextAtlas5M~\cite{wang2025textatlas5m} & \makecell[c]{Text-rich Visuals} & 5M & T2I & \makecell[c]{Real +\\Synthetic} & \Checkmark & \XSolidBrush & Partial \\
    S1-Omni-Image~\cite{li2026s1omniimageunifiedmodelscientific} & \makecell[c]{Scientific Visuals} & 314K & \makecell[c]{T2I + Edit} & Synthetic & \Checkmark & \Checkmark & \Checkmark \\
    \midrule
    \datasetname~(Ours) & \makecell[c]{Multidisciplinary\\Structured Visuals} & 1.2M & \makecell[c]{T2I + Edit} & \makecell[c]{Real + Synthetic\\+ Programmatic} & \Checkmark & \Checkmark & \Checkmark \\
    \bottomrule
    \end{tabular}
}
\caption{\textbf{Comparison with representative image generation and editing datasets.} 
Compared with prior work, \datasetname~focuses on multidisciplinary academic visuals and provides structured supervision with stronger knowledge-intensive characteristics.}
\label{tab:dataset-compare}
\end{table*}

To propel the development of reasoning-informed generation, we target the multidisciplinary domain and aim to propose a dataset tailored for reasoning-informed generation within this field. Compared to natural images, disciplinary illustrations prioritize the correctness of underlying structures and academic knowledge over sheer aesthetic appeal: a small mistake in a symbol, label, graph edge, molecular group, or notation can invalidate the result. Furthermore, the reasoning required here transcends the straightforward rewriting seen in natural image benchmarks such as WISE~\cite{niu2025wiseworldknowledgeinformedsemantic} and RISE~\cite{zhao2025envisioning}, which predominantly rely on general world knowledge or commonsense concepts, where the prompts remain simple descriptions of the visual scene. Instead, disciplinary prompts are embedded with rich academic knowledge, thereby offering a rigorous testbed to elevate models' image generation capabilities under complex reasoning. As summarized in Table~\ref{tab:dataset-compare}, existing structured or text-rich visual datasets usually cover only part of this space, often lacking either multidisciplinary scope, joint T2I/editing supervision, or explicit reasoning and knowledge annotations. This motivates \datasetname, which targets multidisciplinary academic visuals with both generation and editing formats under structured, knowledge-intensive supervision.

We address this gap with \datasetname, which contains 1.2M samples for multidisciplinary text-to-image generation and image editing.
To handle the heterogeneity of academic images, we combine four complementary construction pipelines: structured rendering, OCR-based editing, specialized programmatic synthesis, and large-scale T2I filtering. Together, these pipelines produce 1.2M examples across ten disciplines, covering both T2I and editing supervision for diagrams, charts, formulas, maps, molecular structures, music notation, board states, and other structured academic visuals.

Building on our dataset, we introduce a discipline-informed reasoning-generation model for multidisciplinary visual generation.
We adapt the corresponding Qwen-Image generator for each task under a shared reasoning-generation pipeline. Qwen3-VL-8B~\cite{Qwen3-VL} turns implicit disciplinary requests into explicit reasoning plans, while Qwen-Image-2512 handles text-to-image generation and Qwen-Image-Edit-2511 handles editing~\cite{wu2025qwenimagetechnicalreport}. We first evaluate on discipline-related benchmarks: on GenExam~\cite{wang2025genexam}, our T2I model reaches 51.4 relaxed score, and on GRADE~\cite{liu2026grade}, our editing model reaches 58.7 relaxed score, both ranking first among open-source models. We then evaluate on general reasoning-informed benchmarks, WISE~\cite{niu2025wiseworldknowledgeinformedsemantic} and RISE~\cite{zhao2025envisioning}, observing consistent improvements beyond discipline-specific benchmarks. These results suggest that, beyond scale, exposing models to disciplinary structures, verifiable edits, and explicit knowledge constraints is important for academic visual generation and editing.

Our main contributions are summarized as follows:

\begin{enumerate}[leftmargin=*]
    \item We introduce a scalable multidisciplinary data construction framework that combines structured rendering, OCR-based editing, specialized programmatic synthesis, and large-scale T2I filtering.
    \item We build \datasetname, which to the best of our knowledge is the first million-scale multidisciplinary visual generation dataset.
    \item We adapt open-source image generators with \datasetname and demonstrate strong open-source state-of-the-art results on GenExam and GRADE, with additional generalization gains on WISE and RISE, validating the usefulness of \datasetname for knowledge-grounded visual creation.
\end{enumerate}

\section{Related Work}

\subsection{Models for Image Generation and Editing}

Diffusion-based text-to-image systems have improved fidelity, alignment, and controllability through GLIDE~\cite{nichol2021glide}, Imagen~\cite{saharia2022photorealistic}, latent diffusion~\cite{rombach2022high}, Parti~\cite{yu2022scalingautoregressivemodelscontentrich}, and ControlNet~\cite{zhang2023adding}. Instruction editing progressed from Prompt-to-Prompt~\cite{hertz2022prompttoprompt} and InstructPix2Pix~\cite{brooks2022instructpix2pix} to SmartEdit~\cite{huang2023smarteditexploringcomplexinstructionbased}, ICEdit~\cite{zhang2025icedit}, DreamOmni~\cite{xia2025dreamomniunifiedimagegeneration}, and Qwen-Image~\cite{wu2025qwenimagetechnicalreport}. Unified multimodal models, including Chameleon~\cite{team2024chameleon}, Emu~\cite{sun2024generative}, OmniGen~\cite{xiao2025omnigen}, Janus-Pro~\cite{chen2025januspro}, BAGEL~\cite{deng2025emerging}, and InternVL-U~\cite{internvl-u}, connect understanding with generation. However, these advances remain focused largely on natural images, and capacity alone does not ensure structurally correct academic diagrams; \datasetname adds explicit supervision for academic formats, symbolic constraints, and verifiable transformations.

\subsection{Datasets for Image Generation and Editing}

General T2I datasets span broad image--text corpora (Conceptual 12M~\cite{changpinyo2021cc12m}, WIT~\cite{srinivasan2021wit}, LAION-5B~\cite{schuhmann2022laion5b}) and prompt-enriched resources (DiffusionDB~\cite{wang2022diffusiondb}, PixelProse~\cite{singla2024pixelprose}); editing datasets include InstructPix2Pix~\cite{brooks2022instructpix2pix}, MagicBrush~\cite{zhang2023magicbrush}, UltraEdit~\cite{zhao2024ultraedit}, OmniEdit~\cite{wei2024omniedit}, AnyEdit~\cite{yu2024anyedit}, Pico-Banana-400K~\cite{qian2025picobanana}, and ScaleEdit-12M~\cite{chen2026scaleedit}. These resources emphasize web captions and natural imagery. TextAtlas5M~\cite{wang2025textatlas5m} and StructVisuals~\cite{zhuo2025factuality} move toward text-rich and structured visuals, while GenExam~\cite{wang2025genexam}, GRADE~\cite{liu2026grade}, WISE~\cite{niu2025wiseworldknowledgeinformedsemantic}, and RISE~\cite{zhao2025envisioning} expose discipline- and reasoning-related weaknesses. None provides large-scale paired T2I/editing supervision across many academic domains; \datasetname fills this gap with multidisciplinary, structured, and verifiable data.

\section{Data Construction}

\begin{figure}[t]
  \centering
  \includegraphics[width=\textwidth]{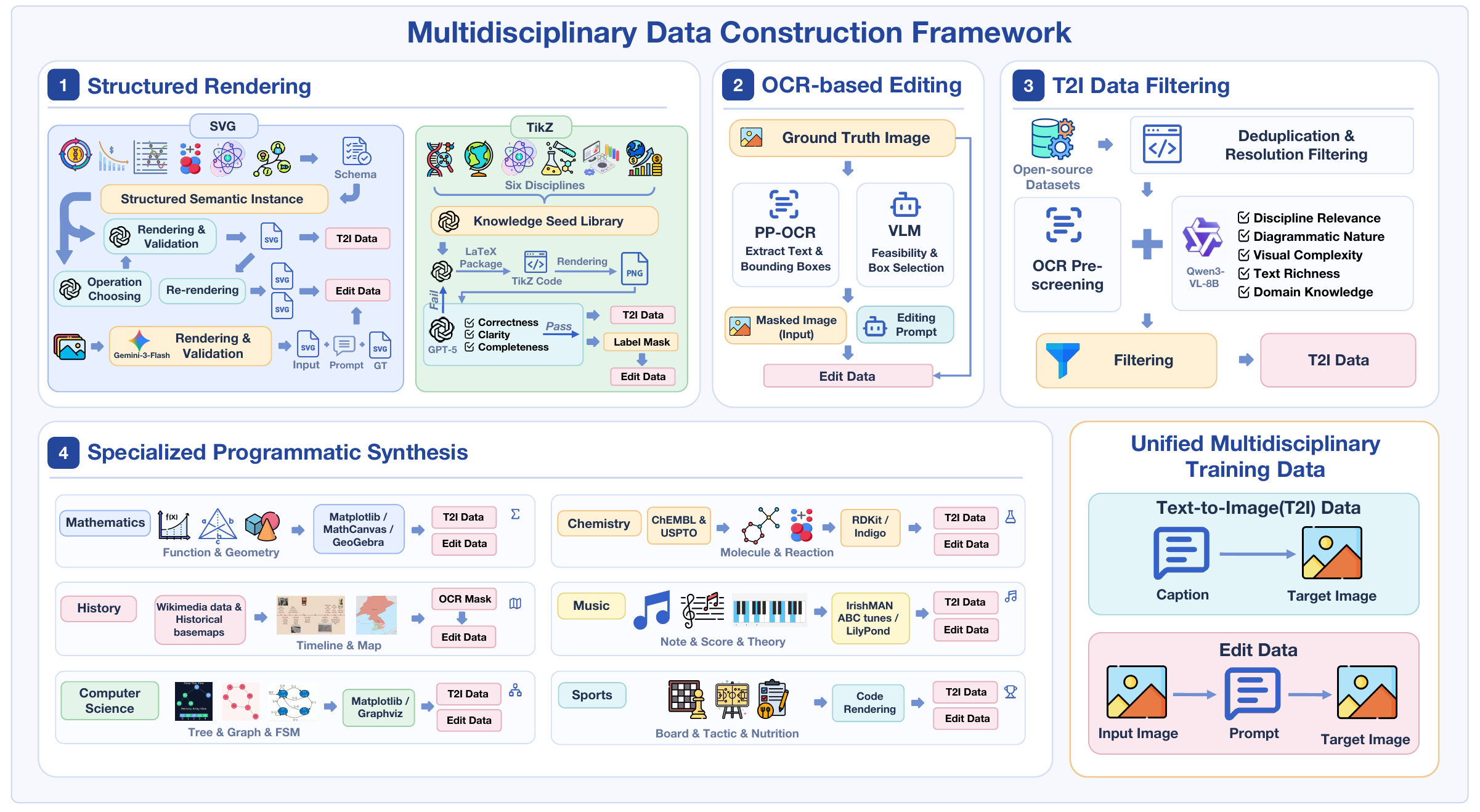}
  \caption{\textbf{Overview of the \datasetname construction framework.} We combine four complementary methods to produce T2I (image and caption) and editing data (image pair and instruction): structured rendering with SVG/TikZ, OCR-based editing, large-scale T2I filtering, and specialized programmatic synthesis.}
  \label{fig:dataset_construction}
\end{figure}

\begin{figure}[htbp]
  \centering
  \includegraphics[width=\textwidth]{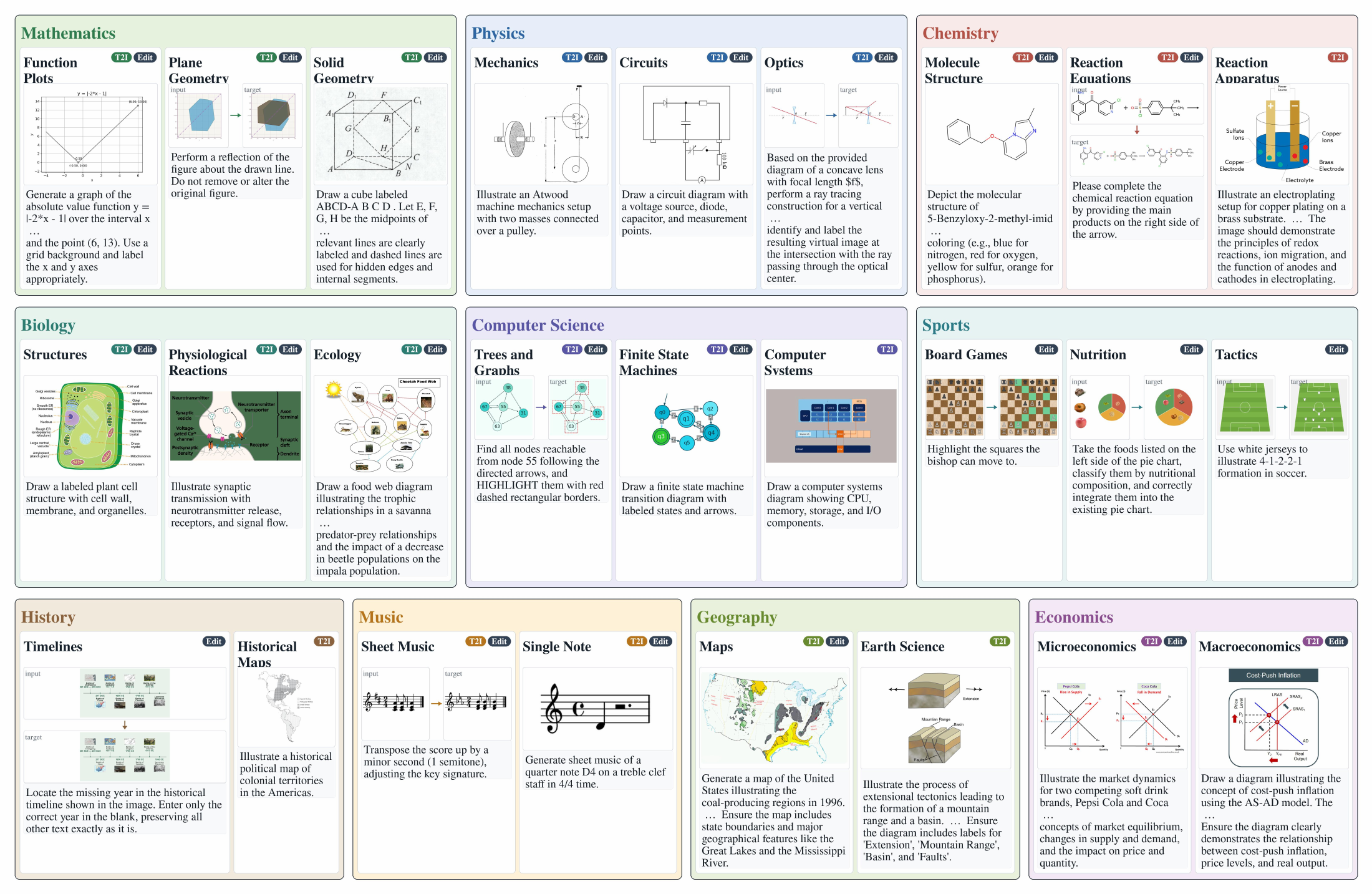}
  \caption{\textbf{Representative examples of each discipline and sub-discipline from \datasetname.} Our multidisciplinary data span over ten disciplines and their fine-grained subdomains. For sub-disciplines that contain both T2I and editing samples, we only display one of them as an example. Long prompts are truncated in the middle.}
  \label{fig:dataset_example}
\end{figure}

To obtain a large-scale multidisciplinary text-to-image (T2I) and image editing dataset, we design the construction framework shown in Figure~\ref{fig:dataset_construction}. The central challenge is that academic images are not a single visual domain. Some are symbolic vector diagrams, some are text-label completion tasks, some require domain-specific engines, and others can only be obtained by filtering noisy open-source educational images. We therefore combine four pipelines: (1) structured data rendering; (2) OCR-based editing; (3) large-scale T2I data filtering; and (4) specialized programmatic synthesis. Each pipeline emits a unified annotation format so that the resulting data can support both generation and editing training.

Figure~\ref{fig:dataset_example} illustrates the resulting data format and visual diversity. A sample is organized by discipline and fine-grained subdomain, and then instantiated as either a T2I example with an image and caption-like prompt or an editing example with an input image, a target image, and a natural-language instruction.

\subsection{Structured Data Rendering}

Collecting multidisciplinary images from existing sources often leads to limited coverage, quality, and resolution. The problem is harder for editing because high-quality paired images are rarely available. Directly using a proprietary image editing model such as Nano Banana Pro~\cite{Nano-Banana-Pro} to create paired data is also expensive and may introduce uncontrolled semantic changes. Many academic diagrams, however, are naturally described by vector primitives: text labels, arrows, curves, axes, equations, symbols, and spatial relations. Motivated by recent progress in SVG and TikZ understanding and generation~\cite{wang2025internsvg,sgpbench,lin2026scientific,belouadi2024automatikz}, we synthesize a large portion of \datasetname through structured vector rendering. Operating on SVG/TikZ code gives us high-resolution images, deterministic re-rendering, and explicit control over semantic edits.

\subsubsection{SVG-Based Data Rendering}

As illustrated in Figure~\ref{fig:dataset_construction}, the SVG branch constructs T2I and editing data for domains such as biology, physics, and economics. Each image is first represented as a structured semantic instance containing domain-specific fields, such as textual nodes, directed relations, curve parameters, reaction components, axis labels, and scientific constraints. The instance is mapped to a diagram template and rendered into SVG through sampled layout, style, and typography. The same structured representation also generates captions, which gives aligned image-caption pairs for T2I training.
For editing data, we apply verified transformations to the structured representation and re-render both source and target images. Examples include removing or restoring labels, shifting curves, changing key annotations, or modifying reaction components.

The SVG branch also supports web image initialization: after extracting a candidate image and its context, Gemini-3-Flash~\cite{Gemini-3-Flash} generates SVG code for both the original and edited diagrams, followed by automatic rendering and filtering. Compared with directly prompting Nano Banana Pro~\cite{Nano-Banana-Pro} to generate the edited output, this SVG-based route reduces the estimated cost from \$0.16 to \$0.03 per sample while preserving stronger geometric consistency.

\subsubsection{TikZ-Based Data Rendering}

The TikZ branch targets biology, geography, physics, chemistry, computer science, and economics, where many diagrams can be expressed as compilable graphics programs. We first build a seed library of 1,346 items covering 17 fine-grained knowledge points across six disciplines. Each seed is expanded into a drawing blueprint and translated into TikZ source. A compiler-in-the-loop repair process handles LaTeX errors, and a VLM judge filters rendered images according to correctness, clarity, and completeness. For editing, samples that contain explicit labels are rewritten by replacing each label with numbered placeholders and recompiling the same source file. This creates fill-in editing pairs that are pixel-aligned except for the annotation tokens.

\subsection{OCR Editing Pipeline}

Many educational figures contain informative labels whose removal and recovery naturally define editing tasks. The OCR editing pipeline turns such figures into paired data without asking an image model to hallucinate a target. First, all text instances and bounding boxes are extracted using PP-OCR v5~\cite{paddleocr3}. Second, the image and OCR results are passed to a VLM that selects regions suitable for masking, rejecting boxes that are too small, visually ambiguous, or semantically unimportant. Finally, selected boxes are masked to form the input image, while the original image serves as the target. The VLM also writes the editing instruction, producing aligned triplets of input image, instruction, and ground-truth image.

\begin{figure}[t]
  \centering
  \includegraphics[width=\linewidth]{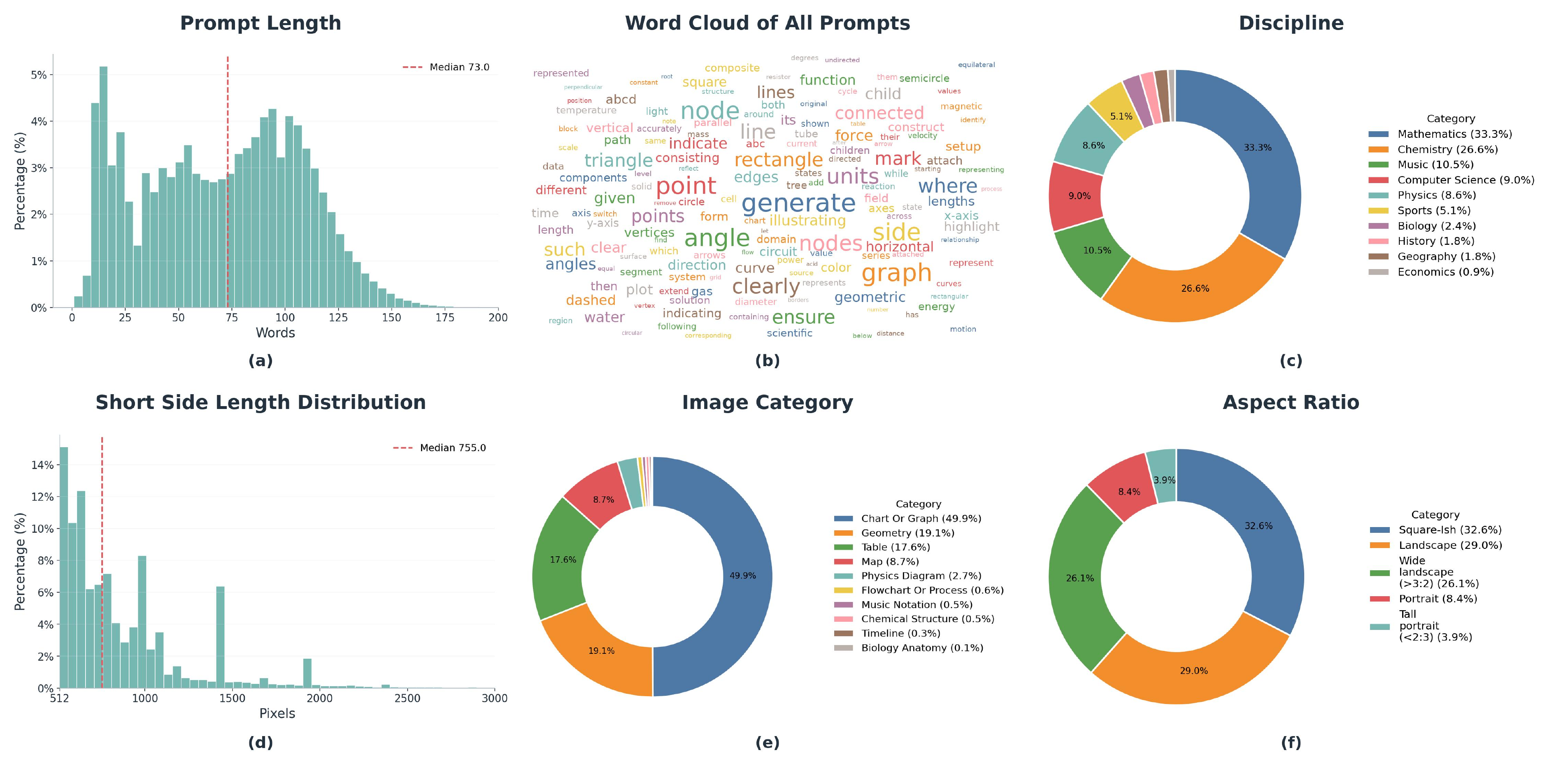}
  \caption{\textbf{Statistics of \datasetname.} The dataset contains long and information-dense prompts, diverse discipline coverage, multiple image categories, and varied resolutions and aspect ratios.}
  \label{fig:statistics}
\end{figure}

\subsection{T2I Data Filtering Pipeline}

Programmatic synthesis offers precise control but cannot cover every real-world academic visual style. We therefore filter large open-source educational image pools for T2I training. The pipeline first applies pHash~\cite{zauner2010phash} deduplication, benchmark-overlap removal, and resolution filtering. It then performs OCR\cite{paddleocr3} pre-screening to remove language-dependent or text-dominant images. Remaining images are scored by Qwen3-VL-8B-Instruct~\cite{Qwen3-VL} along multiple dimensions, including discipline relevance, diagrammatic nature, visual complexity, text richness, and required domain knowledge. Rule-based post-filtering keeps non-photographic, structurally informative, moderately complex images that are suitable for discipline-specific T2I generation. Detailed thresholds and score distributions are provided in Appendix~\ref{sec:appendix}.

\subsection{Specialized Synthesis Pipelines}

Editing data requires paired images with high consistency, which are difficult to directly obtain from open-source datasets and web images. In addition, existing T2I images fail to cover certain domains with sufficient data. We therefore design specialized synthesis pipelines for math, chemistry, computer science, history, music, and sports; their sources and task definitions are summarized in Appendix~\ref{sec:appendix}, Table~\ref{tab:specialized_pipelines}.

\begin{table*}
  \caption{\textbf{Specialized synthesis pipelines.} Programmatic engines complement vector rendering and web filtering by creating paired images with deterministic semantic differences in domains where ordinary web data is sparse or difficult to align. Detailed task definitions are provided in the appendix.}
  \label{tab:specialized_pipelines}
  \footnotesize
  \centering
  \setlength{\tabcolsep}{3pt}
  \renewcommand{\arraystretch}{1.2}
  \begin{tabular}{p{0.08\textwidth}p{0.32\textwidth}p{0.55\textwidth}}
  \toprule
  \textbf{Domain} & \textbf{Sources / Engines} & \textbf{Tasks} \\
  \midrule
  Math & Matplotlib, MathCanvas~\cite{shi2025mathcanvasintrinsicvisualchainofthought}, GeoGebra~\cite{geogebra:website} & Function plots, zeros and poles, tangent lines, derivatives, plane geometry transformations, solid rotations, projections, translations, and symmetry edits. \\
  Chemistry & Molecular and reaction records rendered with RDKit~\cite{rdkit} and Indigo~\cite{Pavlov2011} & Molecule generation from SMILES~\cite{Weininger1988}, reaction-equation completion, major-product rendering, and functional-group insertion with validity checks. \\
  Computer Science & Programmatic graph engines, Matplotlib, Graphviz \textit{neato} & Tree, graph, and finite-state-machine tasks, including traversal paths, BST/heap operations, shortest paths, cycle detection, reachability, and transition completion. \\
  History & Wikimedia Commons~\cite{commons:cat}, historical-basemaps~\cite{historical-basemaps} & Timeline completion and historical map label completion with randomized layout, cropping, and label de-overlap. \\
  Music & IrishMAN ABC tunes~\cite{irishman}, MIDI resources~\cite{lakhmidi,MAESTRO}, LilyPond & Single-note editing, text-to-score generation, sheet-music editing, piano-roll editing, theory-chart editing, and sheet-to-roll conversion. \\
  Sports & Chess/Xiangqi records, Go SGF records, nutrition tables, soccer event data & Board-game move reasoning, legal-move and best-move highlighting, nutrition chart editing, and soccer formation or ball-handler tactical diagrams. \\
  \bottomrule
  \end{tabular}
  \end{table*}

\subsection{Dataset Analysis}

Figure~\ref{fig:statistics} summarizes the current data distribution. Prompts are generally long and information-dense, with a median length of 73 words, reflecting the fact that many samples encode not only visual appearance but also disciplinary constraints. Images are also diverse in category, resolution, and aspect ratio. The distribution contains square-ish diagrams, landscape charts and maps, tall notation-like images, and wide structural visuals, which helps expose models to realistic academic layout variation. At the category level, charts/graphs, geometry, tables, maps, molecular structures, music notation, timelines, and biological anatomy provide complementary visual grammars rather than a single dominant natural-image style.

\section{Experiment}

\begin{table}[t]
\centering
\caption{Evaluation of discipline-informed text-to-image generation ability on GenExam~\cite{wang2025genexam} (Relaxed Score).
}
\setlength{\tabcolsep}{4pt}
\renewcommand{\arraystretch}{1.2}
\resizebox{0.75\textwidth}{!}{%
\begin{tabular}{l|cccccccccc|c}
\toprule
\textbf{Model} & \textbf{Math} & \textbf{Phy} & \textbf{Chem} & \textbf{Bio} & \textbf{Geo} & \textbf{Comp} & \textbf{Eng} & \textbf{Econ} & \textbf{Music} & \textbf{Hist} & \textbf{Overall} \\
\midrule

\multicolumn{12}{l}{\textbf{$\blacktriangledown$ Closed-Source Models}} \\
\midrule

GPT-Image-2~\cite{gptimage2} & 85.2 & 95.6 & \textbf{92.0} & \textbf{97.5} & \textbf{97.6} & \textbf{93.3} & \textbf{96.5} & \textbf{97.7} & 89.1 & 97.1 & \textbf{93.8} \\
Nano Banana Pro~\cite{Nano-Banana-Pro} & \textbf{86.3} & 95.1 & 88.7 & 95.9 & 96.5 & 91.7 & 95.1 & 97.2 & \textbf{91.0} & \textbf{99.9} & 93.7 \\
Nano Banana 2~\cite{Nano-Banana-2} & 87.8 & \textbf{95.7 }& 90.0 & 95.2 & 94.8 & 88.8 & 95.8 & 94.2 & 86.9 & 97.3 & 92.6 \\
Seedream 5.0~\cite{Seedream5-0} & 82.9 & 85.7 & 81.1 & 89.9 & 89.5 & 85.2 & 91.2 & 94.7 & 76.7 & 87.0 & 86.4 \\
GPT-Image-1.5~\cite{GPT-Image-1.5} & 65.8 & 85.4 & 78.1 & 91.9 & 92.5 & 75.8 & 86.4 & 85.5 & 70.8 & 90.9 & 82.3 \\
GPT-Image-1~\cite{GPT-Image-1} & 52.0 & 66.4 & 53.4 & 74.6 & 73.9 & 55.6 & 65.5 & 65.8 & 52.6 & 67.4 & 62.6 \\
FLUX.2 Max~\cite{flux-2-2025} & 49.1 & 63.2 & 54.0 & 74.5 & 76.3 & 56.5 & 68.9 & 61.5 & 47.0 & 68.0 & 61.9 \\
Seedream 4.5~\cite{Seedream4-5} & 44.7 & 63.4 & 48.9 & 75.8 & 67.6 & 57.9 & 69.7 & 67.3 & 38.0 & 55.0 & 58.8 \\

\midrule
\multicolumn{12}{l}{\textbf{$\blacktriangledown$ Open-Source Models}} \\
\midrule

FLUX.2 dev~\cite{flux-2-2025} & 31.6 & 42.7 & 33.2 & 54.8 & 62.6 & 31.1 & 48.9 & 43.6 & 33.4 & 47.5 & 42.9 \\
Qwen-Image-2512~\cite{zhao2026qwenimage20} & 27.9 & 41.3 & 23.2 & 44.4 & 56.6 & 24.1 & 42.9 & 32.3 & 28.3 & 37.0 & 35.8 \\

BAGEL (thinking)~\cite{deng2025emerging} & 11.7 & 13.8 & 11.9 & 15.2 & 28.5 & 6.2 & 10.7 & 6.3 & 14.7 & 16.0 & 13.5 \\
Show-o2-7B~\cite{Show-o2} & 10.8 & 11.9 & 4.8 & 12.8 & 33.3 & 4.7 & 11.8 & 7.0 & 8.8 & 14.5 & 12.0 \\

Ours & \textbf{39.3} & \textbf{51.2} & \textbf{53.7} & \textbf{61.9} & \textbf{63.1} & \textbf{41.8} & \textbf{58.4} & \textbf{60.5} & \textbf{44.0} & \textbf{60.9} & \textbf{51.4} \\

\bottomrule
\end{tabular}
}

\label{tab:exp_genexam}
\end{table}

\subsection{Training Setup}

Based on \datasetname, we adapt the image generation backbone for each task setting under a shared two-stage reasoning-generation design. Given an implicit disciplinary instruction, Qwen3-VL-8B~\cite{Qwen3-VL} first produces an explicit reasoning plan that identifies the relevant discipline knowledge, target visual elements, and required structural changes. The task-specific image generator then synthesizes the final output: Qwen-Image-2512 for T2I generation and Qwen-Image-Edit-2511 for editing~\cite{wu2025qwenimagetechnicalreport}. We use LoRA~\cite{hu2021lora} in each setting to preserve the base model's general visual prior while adapting it to academic diagrams. Both reported generators are optimized with the same DiffSynth training pipeline using FlowMatch SFT loss, dynamic-resolution image loading, and an effective batch size of 8. Detailed hyperparameters are provided in Appendix~\ref{sec:training_details}.

For T2I, we train on the text-to-image portion of \datasetname. For editing, we train on the editing portion of \datasetname. We organize evaluation into two groups. First, we evaluate discipline-related benchmarks, using GenExam~\cite{wang2025genexam} for T2I and GRADE~\cite{liu2026grade} for editing. Second, we evaluate general reasoning-informed benchmarks, using WISE\_Verified~\cite{niu2025wiseworldknowledgeinformedsemantic} for world-knowledge-informed T2I generation and RISEBench~\cite{zhao2025envisioning} for temporal, causal, spatial, and logical reasoning in image editing. We follow the official evaluation protocol of each benchmark. Training details are provided in Appendix~\ref{sec:appendix}.
\subsection{Discipline-Related Benchmarks}

\begin{table*}[t]
  \caption{Evaluation of discipline-informed image editing ability on GRADE~\cite{liu2026grade} (Relaxed Score).}
  \vspace{1mm}
  \centering
  \setlength{\tabcolsep}{4pt}
  \renewcommand{\arraystretch}{1.2}
  \resizebox{0.75\linewidth}{!}{
  \begin{tabular}{lcccccccccc|c}
  \toprule
\textbf{Model} & \textbf{Phy} & \textbf{Sports} & \textbf{Chem} & \textbf{Math} & \textbf{Music} & \textbf{Econ} & \textbf{Hist} & \textbf{Geo} & \textbf{Bio} & \textbf{Comp} & \textbf{Overall} \\
  
  \midrule
  \multicolumn{12}{l}{\textbf{$\blacktriangledown$ Closed-Source Models}} \\
\midrule
GPT-Image-2~\cite{gptimage2}  & \textbf{87.5} & \textbf{79.2} & \textbf{92.9} & \textbf{79.8} & \textbf{84.8} & \textbf{94.5} & \textbf{87.4} & \textbf{85.2} & \textbf{92.5} & \textbf{91.1} & \textbf{87.5}  \\
Nano Banana Pro~\cite{Nano-Banana-Pro} & 84.0 & 75.7 & 85.6 & 75.9 & 82.8 & 89.5 & 81.7 & 80.5 & 89.0 & 90.1 & 82.9 \\
Nano Banana 2~\cite{Nano-Banana-2} & 77.2 & 68.8 & 83.2 & 74.7 & 76.7 & 92.6 & 75.5 & 76.2 & 83.2 & 78.1 & 79.1 \\
Seedream 5.0~\cite{Seedream5-0} & 74.1 & 72.9 & 79.7 & 67.4 & 62.6 & 79.5 & 66.2 & 71.5 & 82.6 & 74.8 & 73.7 \\
GPT-Image-1.5~\cite{GPT-Image-1.5} & 60.8 & 67.8 & 72.9 & 58.5 & 62.3 & 72.1 & 53.8 & 77.1 & 74.2 & 69.0 & 66.5 \\
FLUX.2 Max~\cite{flux-2-2025} & 54.7 & 56.0 & 63.2 & 45.4 & 38.6 & 53.0 & 64.2 & 60.6 & 69.7 & 57.8 & 55.7 \\
GPT-Image-1.0~\cite{GPT-Image-1} & 55.4 & 56.7 & 61.9 & 43.3 & 51.5 & 57.8 & 44.5 & 58.4 & 60.2 & 58.0 & 54.2 \\
Seedream 4.5~\cite{Seedream4-5} & 53.6 & 54.4 & 57.4 & 40.5 & 40.7 & 55.8 & 35.6 & 43.2 & 60.2 & 44.9 & 49.7 \\

\midrule
\multicolumn{12}{l}{\textbf{$\blacktriangledown$ Open-Source Models}} \\
\midrule

FLUX.2 dev~\cite{flux-2-2025} & 34.3 & 54.3 & 34.0 & 37.0 & 34.8 & 32.8 & 40.4 & 42.9 & 37.1 & 36.7 & 37.6 \\
DreamOmni~\cite{xia2025dreamomniunifiedimagegeneration} & 40.6 & 52.7 & 50.6 & 40.1 & 47.4 & 44.9 & 35.5 & 35.9 & 47.0 & 48.9 & 44.3 \\
Step-1X (think)~\cite{liu2025step1x} & 30.9 & 45.0 & 34.0 & 29.3 & 31.7 & 27.5 & 49.8 & 48.7 & 38.1 & 23.7 & 34.3 \\
BAGEL~\cite{deng2025emerging} & 27.7 & 44.3 & 35.4 & 34.3 & 34.4 & 27.2 & 38.5 & 35.2 & 34.0 & 30.7 & 33.7 \\
Qwen-Edit-2511~\cite{zhao2026qwenimage20} & 27.7 & 45.1 & 13.7 & 39.5 & 19.8 & 28.6 & 36.9 & 33.0 & 27.5 & 23.9 & 30.0 \\
OmniGen~\cite{xiao2025omnigen} & 19.5 & 44.0 & 15.7 & 15.3 & 30.0 & 11.5 & 24.8 & 23.4 & 21.2 & 31.8 & 21.1 \\
Ours & \textbf{61.6} & \textbf{68.9} & \textbf{58.2} & \textbf{52.1} & \textbf{46.0} & \textbf{59.1} & \textbf{57.5} & \textbf{66.6} & \textbf{65.0} & \textbf{59.1} & \textbf{58.7} \\

  \bottomrule
  \end{tabular}
  }
  \label{tab:model_performance_domain}
\end{table*}

Table~\ref{tab:exp_genexam} reports GenExam results for discipline-informed T2I generation. Our model reaches 51.4, outperforming Qwen-Image-2512 by 15.6 points and FLUX.2 dev by 8.5 points among open-source systems. Gains span all ten disciplines, with especially large improvements in chemistry, economics, music, computer science, and history, indicating better adherence to molecular, graph, notation, and chart constraints. The gap to leading closed-source models remains substantial.

\begin{table*}[t] 
    \centering 
    \setlength{\tabcolsep}{3pt}
    \renewcommand{\arraystretch}{1.3}

    \caption{Comparison on different image types on GenExam~\cite{wang2025genexam} (Relaxed Score).
    } 
    \label{tab:exp-img_type}
    \vspace{1mm}
    \resizebox{0.7\linewidth}{!}{
    \begin{tabular}{lcccccccc}
    \toprule
    Model & \makecell{Chemical\\Structures} & Diagrams & \makecell{Geometric\\Shapes} & Maps & \makecell{Plots \&\\Charts} & \makecell{Sheet\\Music} & \makecell{Trees \&\\Graphs} & Other \\

    \midrule
  \multicolumn{9}{l}{\textbf{$\blacktriangledown$ Closed-Source Models}} \\
    \midrule
    Nano Banana Pro~\cite{Nano-Banana-Pro} & 85.2 & 94.5 & 84.2 & 97.6 & 96.3 & 90.8 & 89.3 & 89.8 \\ 
    GPT-Image-1.5~\cite{GPT-Image-1.5} & 73.7 & 85.0 & 66.4 & 92.7 & 84.2 & 71.1 & 66.2 & 77.9 \\ 
    FLUX.2 max~\cite{flux-2-2025} & 46.1 & 67.2 & 56.1 & 67.6 & 57.5 & 48.5 & 51.4 & 60.8 \\

    \midrule
    \multicolumn{9}{l}{\textbf{$\blacktriangledown$ Open-Source Models}} \\
      \midrule
    FLUX.2 dev~\cite{flux-2-2025} & 28.5 & 47.7 & 32.2 & 52.1 & 37.0 & 34.4 & 28.8 & 43.3 \\ 
    BAGEL (thinking)~\cite{deng2025emerging} & 10.5 & 13.8 & 13.6 & 22.5 & 7.5 & 15.5 & 8.4 & 9.0 \\ 

    Qwen-Image-2512~\cite{zhao2026qwenimage20} & 19.8 & 40.7 & 30.6 & 41.1 & 31.3 & 28.4 & 24.0 & 28.5 \\ 
    Ours & 36.5 & 57.0 & 35.1 & 46.0 & 53.0 & 30.3 & 30.1 & 75.2 \\

    \bottomrule
    \end{tabular}
    }
    \vspace{3mm}
    \end{table*}

Table~\ref{tab:exp-img_type} further breaks down GenExam performance by image type. Our model improves over Qwen-Image-2512 across all eight visual categories, with the largest gains on Other ($+46.7$), plots and charts ($+21.7$), chemical structures ($+16.7$), and diagrams ($+16.3$). It also outperforms the open-source baselines on six of the eight categories, indicating that \datasetname is especially effective for structured visuals that require precise symbolic elements, layout relations, and domain-specific graphical conventions. The remaining weaknesses are maps and sheet music, where FLUX.2 dev still obtains higher scores, suggesting that geographic layout reasoning and dense music notation may require more targeted data or specialized rendering constraints. Compared with closed-source models, the gap remains substantial on highly structured formats such as maps, sheet music, and trees/graphs, highlighting these image types as useful directions for future data expansion.

Table~\ref{tab:model_performance_domain} reports GRADE results for discipline-informed editing. Our model reaches 58.7, exceeding DreamOmni by 14.4 points and Qwen-edit-2511 by 28.7 points. Gains span physics, chemistry, mathematics, geography, biology, computer science, and history, showing that paired structured data supports targeted edits while preserving surrounding content. The model narrows the open-source gap but still trails GPT-Image-2 and Nano Banana Pro.

\subsection{General Reasoning-Informed Benchmarks}

\begin{table*}[t]
  \caption{Evaluation of world-knowledge-informed text-to-image generation on WISE\_Verified~\cite{niu2025wiseworldknowledgeinformedsemantic}.}
  \vspace{1mm}
  \centering
  \setlength{\tabcolsep}{4pt}
  \renewcommand{\arraystretch}{1.2}
  \resizebox{0.7\linewidth}{!}{
  \begin{tabular}{lcccccc|c}
  \toprule
  \textbf{Model} & \textbf{Culture} & \textbf{Time} & \textbf{Space} & \textbf{Biology} & \textbf{Physics} & \textbf{Chemistry} & \textbf{Overall} \\
  \midrule
  \multicolumn{8}{l}{\textbf{$\blacktriangledown$ Closed-Source Models}} \\
  \midrule
  Nano Banana Pro~\cite{Nano-Banana-Pro} & 0.90 & 0.82 & 0.93 & 0.82 & 0.87 & 0.88 & 0.88 \\
  GPT-Image-1.5~\cite{GPT-Image-1.5} & 0.89 & 0.69 & 0.88 & 0.80 & 0.76 & 0.78 & 0.83 \\
  SenseNova-U1-8B-MoT~\cite{sensenova2026sensenovau1} & 0.69 & 0.60 & 0.74 & 0.63 & 0.72 & 0.69 & 0.68 \\
  DeepGen 1.0~\cite{wang2026deepgen} & 0.65 & 0.41 & 0.72 & 0.39 & 0.59 & 0.45 & 0.57 \\
  \midrule
  \multicolumn{8}{l}{\textbf{$\blacktriangledown$ Open-Source Models}} \\
  \midrule
  BAGEL (thinking)~\cite{deng2025emerging} & 0.78 & 0.63 & 0.57 & 0.38 & 0.55 & 0.51 & 0.63 \\
  FLUX.2-dev~\cite{flux-2-2025} & 0.67 & 0.57 & 0.66 & 0.37 & 0.53 & 0.38 & 0.57 \\
  Qwen-Image-2512~\cite{zhao2026qwenimage20} & 0.60 & 0.48 & 0.60 & 0.35 & 0.49 & 0.26 & 0.50 \\
  BAGEL~\cite{deng2025emerging} & 0.41 & 0.35 & 0.31 & 0.20 & 0.44 & 0.26 & 0.35 \\

  Ours & 0.69 & 0.60 & 0.75 & 0.63 & 0.68 & 0.55 & 0.66 \\
  \bottomrule
  \end{tabular}
  }
  \label{tab:wise_results}
\end{table*}

\begin{table*}[t]
  \caption{Evaluation of reasoning-informed image editing ability on RISEBench~\cite{zhao2025envisioning}.}
  \vspace{1mm}
  \centering
  \setlength{\tabcolsep}{4pt}
  \renewcommand{\arraystretch}{1.2}
  \resizebox{0.55\linewidth}{!}{
  \begin{tabular}{lcccc|c}
  \toprule
  \textbf{Model} & \textbf{Temporal} & \textbf{Causal} & \textbf{Spatial} & \textbf{Logical} & \textbf{Overall} \\
  \midrule
  \multicolumn{6}{l}{\textbf{$\blacktriangledown$ Closed-Source Models}} \\
  \midrule
  GPT-Image-2~\cite{gptimage2} & 45.9 & 66.7 & 50.0 & 34.1 & 49.4 \\
  Nano Banana Pro~\cite{Nano-Banana-Pro} & 43.5 & 63.3 & 48.0 & 37.6 & 48.3 \\
  GPT-Image-1~\cite{GPT-Image-1} & 36.5 & 34.4 & 37.0 & 10.6 & 30.0 \\
  Seedream-4.0~\cite{seedream2025seedream} & 17.6 & 13.3 & 11.0 & 7.1 & 12.2 \\
  \midrule
  \multicolumn{6}{l}{\textbf{$\blacktriangledown$ Open-Source Models}} \\
  \midrule
  Qwen-Image-Edit-2511~\cite{zhao2026qwenimage20} & 21.2 & 18.9 & 31.0 & 4.7 & 19.4 \\
  BAGEL (thinking)~\cite{deng2025emerging} & 11.8 & 27.8 & 21.0 & 1.2 & 15.8 \\
  BAGEL~\cite{deng2025emerging} & 3.5 & 4.4 & 14.0 & 0.0 & 5.8 \\
  Ours & 28.2 & 25.6 & 40.0 & 9.4 & 26.4 \\
  \bottomrule
  \end{tabular}
  }
  \label{tab:rise_results}
\end{table*}

Table~\ref{tab:wise_results} evaluates generalization to WISE\_Verified. Our model obtains 0.66, outperforming BAGEL (w/ CoT) by 0.03, FLUX.2-dev by 0.10, and Qwen-Image-2512 by 0.16. Gains are strongest on spatial, biology, physics, and chemistry categories, consistent with the structured supervision in \datasetname.

Table~\ref{tab:rise_results} reports RISEBench results across temporal, causal, spatial, and logical reasoning. Our model reaches 26.4, outperforming Qwen-Image-Edit-2511 by 7.0 points and BAGEL (thinking) by 10.8 points, with gains over the strongest official closed-source baseline driven mainly by temporal, causal, and spatial reasoning. The weaker logical score suggests room for stronger planning or targeted paired data, while the overall result supports transfer from structured academic editing pairs.

\subsection{Ablation Study}

Table~\ref{tab:ablation} shows the cumulative contribution of prompt enhancement and each data-construction component. For T2I, prompt enhancement raises GenExam from 35.8 to 41.0 and WISE from 0.50 to 0.59; T2I filtering provides the largest subsequent gain, followed by structured rendering, while OCR and specialized data add complementary coverage. For editing, prompt enhancement gives the largest initial improvement, and structured rendering plus OCR supervision drive the main GRADE gains. T2I filtering and specialized programmatic data provide smaller but useful improvements on RISE, indicating complementary transfer from broader data diversity.

\subsection{Qualitative Results}

\begin{figure}[t]
  \centering
  \includegraphics[width=\linewidth]{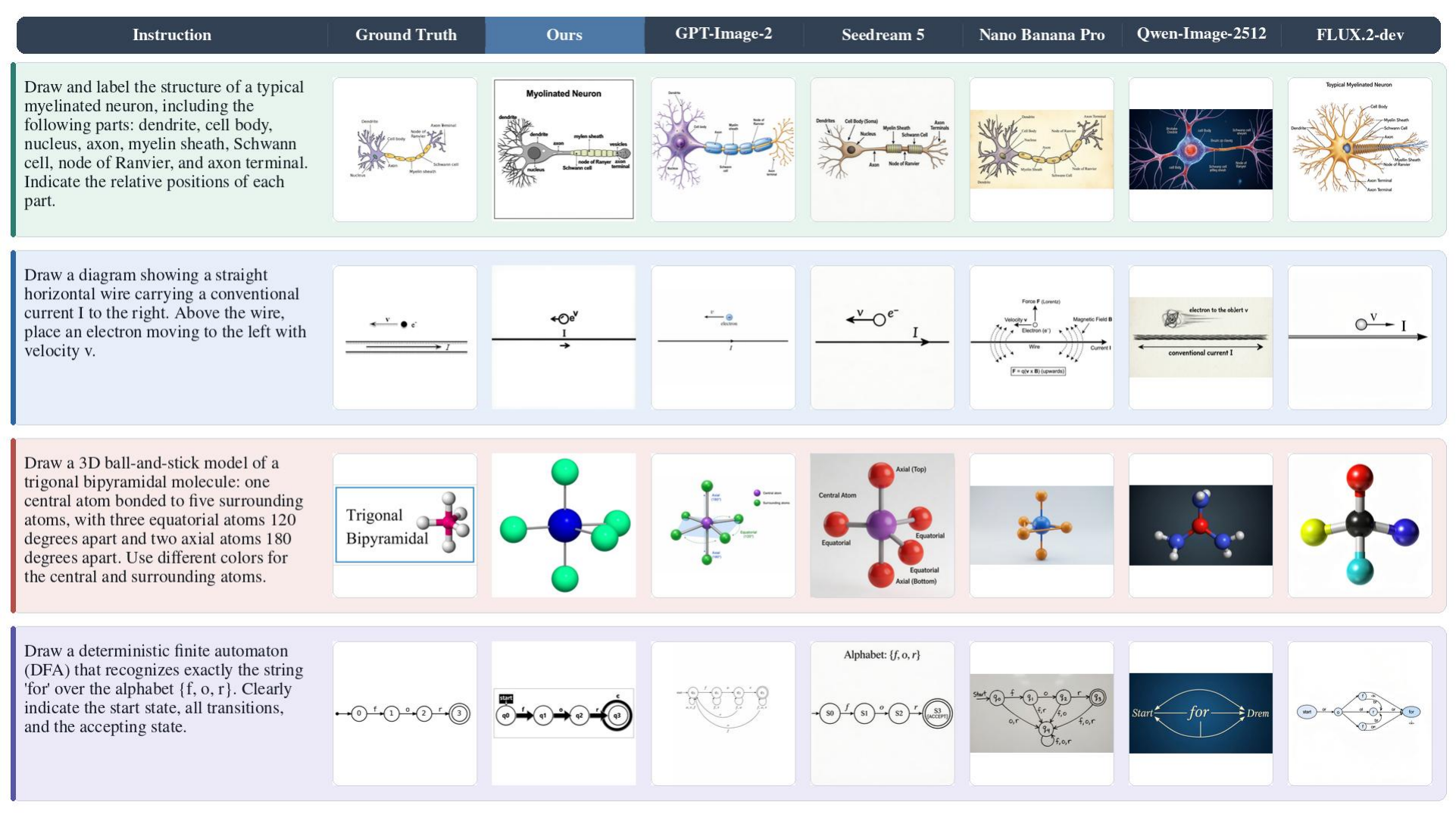}
  \vspace{-5mm}
  \caption{Qualitative text-to-image examples from our model and baselines on GenExam.}
  \label{fig:gen_examples}
\end{figure}

\begin{figure}[t]
  \centering
  \includegraphics[width=\linewidth]{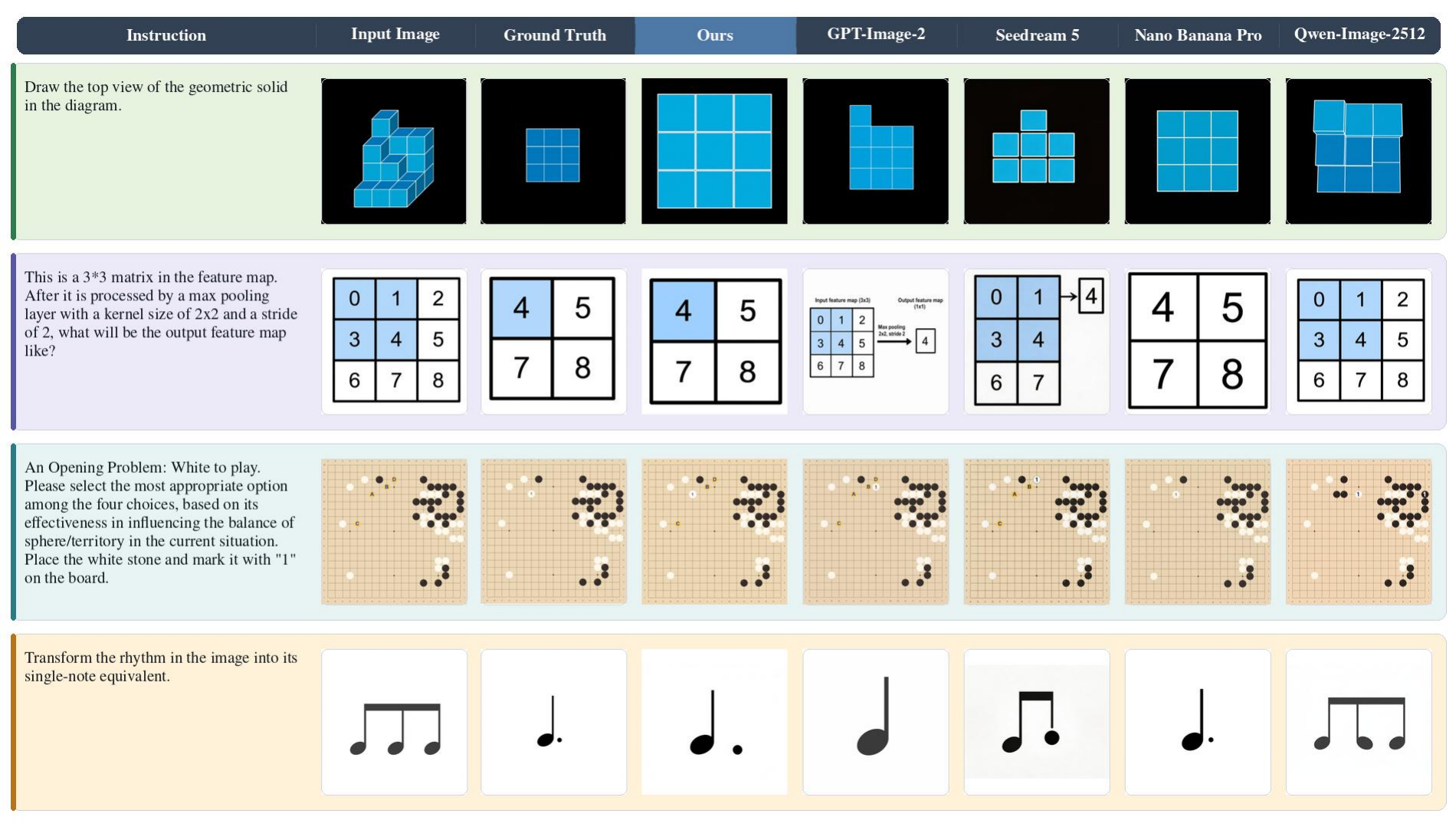}
  \caption{Qualitative editing examples from our model and baselines on GRADE.}
  \label{fig:grade_examples}
\end{figure}

Figure~\ref{fig:gen_examples} presents qualitative GenExam comparisons. Across biology, physics, chemistry, and computer science, our model better preserves constraints such as neuron-part labels, current/electron directions, molecular geometry, and DFA states and transitions. Baselines often miss labels or violate requested spatial and symbolic relations, supporting the quantitative gains from \datasetname. Qualitative editing comparisons are provided in Figure~\ref{fig:grade_examples}.

\begin{table*}[t]
  \centering
  \caption{Ablation study with cumulative data-construction components.}
  \label{tab:ablation}
  \vspace{1mm}
  \begin{minipage}[t]{0.47\textwidth}
    \centering
    \footnotesize
    \setlength{\tabcolsep}{2pt}
    \begin{tabular}{lcc}
      \toprule
      Variant & GenExam & WISE \\
      \midrule
      \makecell[l]{Baseline\\(Qwen-Image-2512)} & 35.8 & 0.50 \\
      + prompt enhancement & 41.0 & 0.59 \\
      + T2I Filtering & 47.4 & 0.62 \\
      + Structured Rendering & 50.2 & 0.64 \\
      + OCR-based Editing & 50.5 & 0.64 \\
      + Specialized Programmatics & 51.4 & 0.66 \\
      \bottomrule
    \end{tabular}
  \end{minipage}
  \begin{minipage}[t]{0.47\textwidth}
    \centering
    \footnotesize
    \setlength{\tabcolsep}{2pt}
    \begin{tabular}{lcc}
      \toprule
      Variant & GRADE & RISE \\
      \midrule
      \makecell[l]{Baseline\\(Qwen-Image-Edit-2511)} & 30.0 & 19.4 \\
      + prompt enhancement & 41.2 & 23.5 \\
      + T2I Filtering & 41.9 & 23.8 \\
      + Structured Rendering & 50.5 & 24.9 \\
      + OCR-based Editing & 54.5 & 24.7 \\
      + Specialized Programmatic & 58.7 & 26.4 \\
      \bottomrule
    \end{tabular}
  \end{minipage}
\end{table*}

Figure~\ref{fig:grade_examples} further shows qualitative editing comparisons on representative GRADE instructions. The examples require targeted modifications while preserving the surrounding academic structure, such as editing labels, symbols, graphical relations, or domain-specific components. Compared with baselines, our model more reliably performs the requested edit without disrupting unrelated visual elements, which supports the GRADE results in Table~\ref{tab:model_performance_domain} and highlights the value of paired structured supervision for discipline-informed image editing.

\subsection{Discussion}

Together, the results show that structured multidisciplinary data benefits both discipline-specific and general reasoning-informed generation. Improvements are not confined to the most represented domains, suggesting transfer through shared visual primitives such as labels, arrows, plots, graphs, and symbolic notation. WISE improves most on spatial and scientific prompts, while RISE benefits especially on temporal, causal, and spatial edits; its weaker logical score indicates that abstract reasoning remains challenging. Overall, the results show that paired structured supervision improves targeted editing and broader transfer, supporting verifiable structure as a path toward more reliable academic visuals.

\section{Conclusion}

We present \datasetname, a million-scale dataset for text-to-image generation and editing of structured multidisciplinary academic visuals. Four pipelines provide broad coverage and explicit structural supervision, including structured rendering, OCR-based editing, specialized programmatic synthesis, and large-scale filtering. Models adapted with \datasetname achieve state-of-the-art open-source results on GenExam and GRADE and strong transfer to WISE and RISE, showing that verifiable academic structure can improve both visual plausibility and knowledge correctness.

\bibliographystyle{plain}
\bibliography{custom}

@misc{commons:cat,
  author    = {Wikimedia Commons},
  title     = {Wikimedia Commons},
  year      = {2026},
  url       = {https://commons.wikimedia.org/wiki/Main_Page},
  note      = {Online; accessed 14-May-2026}
}

@misc{historical-basemaps,
  author       = {aourednik},
  title        = {historical-basemaps},
  year         = {2026},
  url          = {https://github.com/aourednik/historical-basemaps},
  note         = {Online; accessed 14-May-2026}
}

@misc{shi2025mathcanvasintrinsicvisualchainofthought,
      title={MathCanvas: Intrinsic Visual Chain-of-Thought for Multimodal Mathematical Reasoning}, 
      author={Weikang Shi and Aldrich Yu and Rongyao Fang and Houxing Ren and Ke Wang and Aojun Zhou and Changyao Tian and Xinyu Fu and Yuxuan Hu and Zimu Lu and Linjiang Huang and Si Liu and Rui Liu and Hongsheng Li},
      year={2025},
      eprint={2510.14958},
      archivePrefix={arXiv},
      primaryClass={cs.CV},
      url={https://arxiv.org/abs/2510.14958}, 
}

@misc{geogebra:website,
  author    = {GeoGebra Team},
  title     = {{GeoGebra}},
  year      = {2026},
  url       = {https://www.geogebra.org},
  note      = {Online; accessed 14-May-2026}
}

@article{internvl-u,
  title={Internvl-u: Democratizing unified multimodal models for understanding, reasoning, generation and editing},
  author={Tian, Changyao and Yang, Danni and Chen, Guanzhou and Cui, Erfei and Wang, Zhaokai and Duan, Yuchen and Yin, Penghao and Chen, Sitao and Yang, Ganlin and Liu, Mingxin and others},
  journal={arXiv preprint arXiv:2603.09877},
  year={2026}
}

@article{sensenova2026sensenovau1,
  title        = {SenseNova-U1: Unifying Multimodal Understanding and Generation with NEO-unify Architecture},
  author       = {Diao, Haiwen and Wu, Penghao and Deng, Hanming and Wang, Jiahao and Bai, Shihao and Wu, Silei and Fan, Weichen and Ye, Wenjie and Tong, Wenwen and Fan, Xiangyu and others},
  journal      = {arXiv preprint arXiv:2605.12500},
  year         = {2026}
}

@article{liu2026grade,
  title={GRADE: Benchmarking Discipline-Informed Reasoning in Image Editing},
  author={Liu, Mingxin and Fan, Ziqian and Wang, Zhaokai and Gu, Leyao and Zhu, Zirun and He, Yiguo and Yang, Yuchen and Tian, Changyao and Zhao, Xiangyu and Liao, Ning and others},
  journal={arXiv preprint arXiv:2603.12264},
  year={2026}
}

@inproceedings{zhang2023adding,
  title={Adding conditional control to text-to-image diffusion models},
  author={Zhang, Lvmin and Rao, Anyi and Agrawala, Maneesh},
  booktitle={Proceedings of the IEEE/CVF international conference on computer vision},
  pages={3836--3847},
  year={2023}
}

@article{saharia2022photorealistic,
  title={Photorealistic text-to-image diffusion models with deep language understanding},
  author={Saharia, Chitwan and Chan, William and Saxena, Saurabh and Li, Lala and Whang, Jay and Denton, Emily L and Ghasemipour, Kamyar and Gontijo Lopes, Raphael and Karagol Ayan, Burcu and Salimans, Tim and others},
  journal={Advances in neural information processing systems},
  volume={35},
  pages={36479--36494},
  year={2022}
}

@misc{nichol2021glide,
  title={GLIDE: Towards Photorealistic Image Generation and Editing with Text-Guided Diffusion Models},
  author={Nichol, Alex and Dhariwal, Prafulla and Ramesh, Aditya and Shyam, Pranav and Mishkin, Pamela and McGrew, Bob and Sutskever, Ilya and Chen, Mark},
  year={2021},
  eprint={2112.10741},
  archivePrefix={arXiv},
  primaryClass={cs.CV},
  url={https://arxiv.org/abs/2112.10741}
}

@inproceedings{rombach2022high,
  title={High-Resolution Image Synthesis with Latent Diffusion Models},
  author={Rombach, Robin and Blattmann, Andreas and Lorenz, Dominik and Esser, Patrick and Ommer, Bj{\"o}rn},
  booktitle={Proceedings of the IEEE/CVF Conference on Computer Vision and Pattern Recognition},
  pages={10684--10695},
  year={2022}
}

@misc{hertz2022prompttoprompt,
  title={Prompt-to-Prompt Image Editing with Cross Attention Control},
  author={Hertz, Amir and Mokady, Ron and Tenenbaum, Jay and Aberman, Kfir and Pritch, Yael and Cohen-Or, Daniel},
  year={2022},
  eprint={2208.01626},
  archivePrefix={arXiv},
  primaryClass={cs.CV},
  url={https://arxiv.org/abs/2208.01626}
}

@misc{brooks2022instructpix2pix,
  title={InstructPix2Pix: Learning to Follow Image Editing Instructions},
  author={Brooks, Tim and Holynski, Aleksander and Efros, Alexei A.},
  year={2022},
  eprint={2211.09800},
  archivePrefix={arXiv},
  primaryClass={cs.CV},
  url={https://arxiv.org/abs/2211.09800}
}

@misc{changpinyo2021cc12m,
  title={Conceptual 12M: Pushing Web-Scale Image-Text Pre-Training To Recognize Long-Tail Visual Concepts},
  author={Changpinyo, Soravit and Sharma, Piyush and Ding, Nan and Soricut, Radu},
  year={2021},
  eprint={2102.08981},
  archivePrefix={arXiv},
  primaryClass={cs.CV},
  url={https://arxiv.org/abs/2102.08981}
}

@misc{srinivasan2021wit,
  title={WIT: Wikipedia-based Image Text Dataset for Multimodal Multilingual Machine Learning},
  author={Srinivasan, Krishna and Raman, Karthik and Chen, Jiecao and Bendersky, Michael and Najork, Marc},
  year={2021},
  eprint={2103.01913},
  archivePrefix={arXiv},
  primaryClass={cs.CL},
  url={https://arxiv.org/abs/2103.01913}
}

@misc{schuhmann2022laion5b,
  title={LAION-5B: An open large-scale dataset for training next generation image-text models},
  author={Schuhmann, Christoph and Beaumont, Romain and Vencu, Richard and Gordon, Cade and Wightman, Ross and Cherti, Mehdi and Coombes, Theo and Katta, Aarush and Mullis, Clayton and Wortsman, Mitchell and Schramowski, Patrick and Kundurthy, Srivatsa and Crowson, Katherine and Schmidt, Ludwig and Kaczmarczyk, Robert and Jitsev, Jenia},
  year={2022},
  eprint={2210.08402},
  archivePrefix={arXiv},
  primaryClass={cs.CV},
  url={https://arxiv.org/abs/2210.08402}
}

@misc{wang2022diffusiondb,
  title={DiffusionDB: A Large-scale Prompt Gallery Dataset for Text-to-Image Generative Models},
  author={Wang, Zijie J. and Montoya, Evan and Munechika, David and Yang, Haoyang and Hoover, Benjamin and Chau, Duen Horng},
  year={2022},
  eprint={2210.14896},
  archivePrefix={arXiv},
  primaryClass={cs.HC},
  url={https://arxiv.org/abs/2210.14896}
}

@misc{singla2024pixelprose,
  title={From Pixels to Prose: A Large Dataset of Dense Image Captions},
  author={Singla, Vasu and Yue, Kaiyu and Paul, Sukriti and Shirkavand, Reza and Jayawardhana, Mayuka and Ganjdanesh, Alireza and Huang, Heng and Bhatele, Abhinav and Somepalli, Gowthami and Goldstein, Tom},
  year={2024},
  eprint={2406.10328},
  archivePrefix={arXiv},
  primaryClass={cs.CV},
  url={https://arxiv.org/abs/2406.10328}
}

@misc{wang2025textatlas5m,
  title={TextAtlas5M: A Large-scale Dataset for Dense Text Image Generation},
  author={Wang, Alex Jinpeng and Mao, Dongxing and Zhang, Jiawei and Han, Weiming and Dong, Zhuobai and Li, Linjie and Lin, Yiqi and Yang, Zhengyuan and Qin, Libo and Zhang, Fuwei and Wang, Lijuan and Li, Min},
  year={2025},
  eprint={2502.07870},
  archivePrefix={arXiv},
  primaryClass={cs.CV},
  url={https://arxiv.org/abs/2502.07870}
}

@misc{zhang2023magicbrush,
  title={MagicBrush: A Manually Annotated Dataset for Instruction-Guided Image Editing},
  author={Zhang, Kai and Mo, Lingbo and Chen, Wenhu and Sun, Huan and Su, Yu},
  year={2023},
  eprint={2306.10012},
  archivePrefix={arXiv},
  primaryClass={cs.CV},
  url={https://arxiv.org/abs/2306.10012}
}

@misc{zhao2024ultraedit,
  title={UltraEdit: Instruction-based Fine-Grained Image Editing at Scale},
  author={Zhao, Haozhe and Ma, Xiaojian and Chen, Liang and Si, Shuzheng and Wu, Rujie and An, Kaikai and Yu, Peiyu and Zhang, Minjia and Li, Qing and Chang, Baobao},
  year={2024},
  eprint={2407.05282},
  archivePrefix={arXiv},
  primaryClass={cs.CV},
  url={https://arxiv.org/abs/2407.05282}
}

@misc{wei2024omniedit,
  title={OmniEdit: Building Image Editing Generalist Models Through Specialist Supervision},
  author={Wei, Cong and Xiong, Zheyang and Ren, Weiming and Du, Xinrun and Zhang, Ge and Chen, Wenhu},
  year={2024},
  eprint={2411.07199},
  archivePrefix={arXiv},
  primaryClass={cs.CV},
  url={https://arxiv.org/abs/2411.07199}
}

@misc{yu2024anyedit,
  title={AnyEdit: Mastering Unified High-Quality Image Editing for Any Idea},
  author={Yu, Qifan and Chow, Wei and Yue, Zhongqi and Pan, Kaihang and Wu, Yang and Wan, Xiaoyang and Li, Juncheng and Tang, Siliang and Zhang, Hanwang and Zhuang, Yueting},
  year={2024},
  eprint={2411.15738},
  archivePrefix={arXiv},
  primaryClass={cs.CV},
  url={https://arxiv.org/abs/2411.15738}
}

@misc{qian2025picobanana,
  title={Pico-Banana-400K: A Large-Scale Dataset for Text-Guided Image Editing},
  author={Qian, Yusu and Bocek-Rivele, Eli and Song, Liangchen and Tong, Jialing and Yang, Yinfei and Lu, Jiasen and Hu, Wenze and Gan, Zhe},
  year={2025},
  eprint={2510.19808},
  archivePrefix={arXiv},
  primaryClass={cs.CV},
  url={https://arxiv.org/abs/2510.19808}
}

@article{wang2026deepgen,
  title={DeepGen 1.0: A Lightweight Unified Multimodal Model for Advancing Image Generation and Editing},
  author={Wang, Dianyi and Li, Ruihang and Han, Feng and Ma, Chaofan and Song, Wei and Wang, Siyuan and Wang, Yibin and Xin, Yi and Liu, Hongjian and Zhang, Zhixiong and others},
  journal={arXiv preprint arXiv:2602.12205},
  year={2026}
}

@misc{chen2026scaleedit,
  title={ScaleEdit-12M: Scaling Open-Source Image Editing Data Generation via Multi-Agent Framework},
  author={Chen, Guanzhou and Cui, Erfei and Tian, Changyao and Yang, Danni and Yang, Ganlin and Qiao, Yu and Li, Hongsheng and Luo, Gen and Zhang, Hongjie},
  year={2026},
  eprint={2603.20644},
  archivePrefix={arXiv},
  primaryClass={cs.CV},
  url={https://arxiv.org/abs/2603.20644}
}

@misc{zhuo2025factuality,
  title={Factuality Matters: When Image Generation and Editing Meet Structured Visuals},
  author={Zhuo, Le and Han, Songhao and Pu, Yuandong and Qiu, Boxiang and Paul, Sayak and Liao, Yue and Liu, Yihao and Shao, Jie and Chen, Xi and Liu, Si and Li, Hongsheng},
  year={2025},
  eprint={2510.05091},
  archivePrefix={arXiv},
  primaryClass={cs.CV},
  url={https://arxiv.org/abs/2510.05091}
}

@inproceedings{hu2021lora,
  title={LoRA: Low-Rank Adaptation of Large Language Models},
  author={Hu, Edward J. and Shen, Yelong and Wallis, Phillip and Allen-Zhu, Zeyuan and Li, Yuanzhi and Wang, Shean and Wang, Lu and Chen, Weizhu},
  booktitle={International Conference on Learning Representations},
  year={2022},
  url={https://arxiv.org/abs/2106.09685}
}

@article{team2024chameleon,
  title={Chameleon: Mixed-modal early-fusion foundation models, 2024},
  author={Team, Chameleon},
  journal={URL https://arxiv. org/abs/2405.09818},
  volume={9},
  number={8},
  year={2024}
}

@mastersthesis{zauner2010phash,
  title={Implementation and Benchmarking of Perceptual Image Hash Functions},
  author={Zauner, Christoph},
  school={University of Applied Sciences Hagenberg},
  year={2010}
}

@article{Weininger1988,
  author = {Weininger, David},
  title = {SMILES, a Chemical Language and Information System. 1. Introduction to Methodology and Encoding Rules},
  journal = {Journal of Chemical Information and Computer Sciences},
  year = {1988},
  volume = {28},
  number = {1},
  pages = {31--36},
  doi = {10.1021/ci00057a005},
  url = {https://doi.org/10.1021/ci00057a005}
}

@misc{rdkit,
  title={RDKit: Open-source cheminformatics},
  author={Landrum, Greg},
  year={2025},
  url={https://www.rdkit.org}
}

@article{Pavlov2011,
  author = {Pavlov, D. and Rybalkin, M. and Karulin, B. and
            Kozhevnikov, M. and Savelyev, A. and Churinov, A.},
  title = {Indigo: universal cheminformatics API},
  journal = {Journal of Cheminformatics},
  year = {2011},
  volume = {3},
  number = {1},
  pages = {P4},
  doi = {10.1186/1758-2946-3-S1-P4},
  url = {https://doi.org/10.1186/1758-2946-3-S1-P4}
}

@article{Schneider2016,
  author = {Schneider, Nadine and Stiefl, Nikolaus and Landrum, Gregory A.},
  title = {What's What: The (Nearly) Definitive Guide to Reaction Role Assignment},
  journal = {Journal of Chemical Information and Modeling},
  year = {2016},
  volume = {56},
  number = {12},
  pages = {2336--2346},
  doi = {10.1021/acs.jcim.6b00564},
  url = {https://doi.org/10.1021/acs.jcim.6b00564}
}

@misc{peng2025bizgenadvancingarticlelevelvisual,
      title={BizGen: Advancing Article-level Visual Text Rendering for Infographics Generation}, 
      author={Yuyang Peng and Shishi Xiao and Keming Wu and Qisheng Liao and Bohan Chen and Kevin Lin and Danqing Huang and Ji Li and Yuhui Yuan},
      year={2025},
      eprint={2503.20672},
      archivePrefix={arXiv},
      primaryClass={cs.CV},
      url={https://arxiv.org/abs/2503.20672}, 
}

@misc{kuchar2025vectoreditsdatasetbenchmarkinstructionbased,
      title={VectorEdits: A Dataset and Benchmark for Instruction-Based Editing of Vector Graphics}, 
      author={Josef Kuchar and Marek Kadlcik and Michal Spiegel and Michal stefanik},
      year={2025},
      eprint={2506.15903},
      archivePrefix={arXiv},
      primaryClass={cs.LG},
      url={https://arxiv.org/abs/2506.15903}, 
}

@misc{li2026s1omniimageunifiedmodelscientific,
      title={S1-Omni-Image: A Unified Model for Scientific Image Understanding, Generation, and Editing}, 
      author={Qingxiao Li and Zikai Wang and Qingli Wang and Nan Xu},
      year={2026},
      eprint={2606.24441},
      archivePrefix={arXiv},
      primaryClass={cs.CV},
      url={https://arxiv.org/abs/2606.24441}, 
}

@article{10.1093/nar/gkad1004,
    author = {Zdrazil, Barbara and Felix, Eloy and Hunter, Fiona and Manners, Emma J and Blackshaw, James and Corbett, Sybilla and de Veij, Marleen and Ioannidis, Harris and Lopez, David Mendez and Mosquera, Juan F and Magarinos, Maria Paula and Bosc, Nicolas and Arcila, Ricardo and Kiziloren, Tevfik and Gaulton, Anna and Bento, A Patricia and Adasme, Melissa F and Monecke, Peter and Landrum, Gregory A and Leach, Andrew R},
    title = {The ChEMBL Database in 2023: a drug discovery platform spanning multiple bioactivity data types and time periods},
    journal = {Nucleic Acids Research},
    volume = {52},
    number = {D1},
    pages = {D1180-D1192},
    year = {2024},
    month = {01},
    issn = {0305-1048},
    doi = {10.1093/nar/gkad1004},
    url = {https://doi.org/10.1093/nar/gkad1004},
    eprint = {https://academic.oup.com/nar/article-pdf/52/D1/D1180/55040046/gkad1004.pdf},
}

@article{deng2025emerging,
  title={Emerging properties in unified multimodal pretraining},
  author={Deng, Chaorui and Zhu, Deyao and Li, Kunchang and Gou, Chenhui and Li, Feng and Wang, Zeyu and Zhong, Shu and Yu, Weihao and Nie, Xiaonan and Song, Ziang and others},
  journal={arXiv preprint arXiv:2505.14683},
  year={2025}
}

@article{liu2025step1x,
  title={Step1x-edit: A practical framework for general image editing},
  author={Liu, Shiyu and Han, Yucheng and Xing, Peng and Yin, Fukun and Wang, Rui and Cheng, Wei and Liao, Jiaqi and Wang, Yingming and Fu, Honghao and Han, Chunrui and others},
  journal={arXiv preprint arXiv:2504.17761},
  year={2025}
}

@inproceedings{xiao2025omnigen,
  title={Omnigen: Unified image generation},
  author={Xiao, Shitao and Wang, Yueze and Zhou, Junjie and Yuan, Huaying and Xing, Xingrun and Yan, Ruiran and Li, Chaofan and Wang, Shuting and Huang, Tiejun and Liu, Zheng},
  booktitle={Proceedings of the Computer Vision and Pattern Recognition Conference},
  pages={13294--13304},
  year={2025}
}

@misc{chen2025januspro,
  title={Janus-Pro: Unified Multimodal Understanding and Generation with Data and Model Scaling},
  author={Chen, Xiaokang and Wu, Zhiyu and Liu, Xingchao and Pan, Zizheng and Liu, Wen and Xie, Zhenda and Yu, Xingkai and Ruan, Chong},
  year={2025},
  eprint={2501.17811},
  archivePrefix={arXiv},
  primaryClass={cs.AI},
  url={https://arxiv.org/abs/2501.17811}
}

@inproceedings{sun2024generative,
  title={Generative multimodal models are in-context learners},
  author={Sun, Quan and Cui, Yufeng and Zhang, Xiaosong and Zhang, Fan and Yu, Qiying and Wang, Yueze and Rao, Yongming and Liu, Jingjing and Huang, Tiejun and Wang, Xinlong},
  booktitle={Proceedings of the IEEE/CVF Conference on Computer Vision and Pattern Recognition},
  pages={14398--14409},
  year={2024}
}

@misc{wu2025qwenimagetechnicalreport,
      title={Qwen-Image Technical Report}, 
      author={Chenfei Wu and Jiahao Li and Jingren Zhou and Junyang Lin and Kaiyuan Gao and Kun Yan and Sheng-ming Yin and Shuai Bai and Xiao Xu and Yilei Chen and Yuxiang Chen and Zecheng Tang and Zekai Zhang and Zhengyi Wang and An Yang and Bowen Yu and Chen Cheng and Dayiheng Liu and Deqing Li and Hang Zhang and Hao Meng and Hu Wei and Jingyuan Ni and Kai Chen and Kuan Cao and Liang Peng and Lin Qu and Minggang Wu and Peng Wang and Shuting Yu and Tingkun Wen and Wensen Feng and Xiaoxiao Xu and Yi Wang and Yichang Zhang and Yongqiang Zhu and Yujia Wu and Yuxuan Cai and Zenan Liu},
      year={2025},
      eprint={2508.02324},
      archivePrefix={arXiv},
      primaryClass={cs.CV},
      url={https://arxiv.org/abs/2508.02324}, 
}

@misc{zhao2026qwenimage20,
  title={Qwen-Image-2.0 Technical Report},
  author={Zhao, Bing and Wu, Chenfei and Li, Deqing and Meng, Hao and Li, Jiahao and Zhang, Jie and Zhou, Jingren and Lin, Junyang and Gao, Kaiyuan and Cao, Kuan and others},
  year={2026},
  eprint={2605.10730},
  archivePrefix={arXiv},
  primaryClass={cs.CV},
  url={https://arxiv.org/abs/2605.10730}
}

@inproceedings{zhang2025icedit,
  title     = {In-Context Edit: Enabling Instructional Image Editing with In-Context Generation in Large-Scale Diffusion Transformers},
  author    = {Zhang, Zechuan and Xie, Ji and Lu, Yu and Yang, Zongxin and Yang, Yi},
  booktitle = {Advances in Neural Information Processing Systems (NeurIPS)},
  year      = {2025},
  note      = {arXiv:2504.20690}
}

@misc{flux-2-2025,
    author={Black Forest Labs},
    title={{FLUX.2: Frontier Visual Intelligence}},
    year={2025},
    howpublished={\url{https://bfl.ai/blog/flux-2}},
}

@article{singh2025openai,
  title={Openai gpt-5 system card},
  author={Singh, Aaditya and Fry, Adam and Perelman, Adam and Tart, Adam and Ganesh, Adi and El-Kishky, Ahmed and McLaughlin, Aidan and Low, Aiden and Ostrow, AJ and Ananthram, Akhila and others},
  journal={arXiv preprint arXiv:2601.03267},
  year={2025}
}

@article{seedream2025seedream,
  title={Seedream 4.0: Toward next-generation multimodal image generation},
  author={Seedream, Team and Chen, Yunpeng and Gao, Yu and Gong, Lixue and Guo, Meng and Guo, Qiushan and Guo, Zhiyao and Hou, Xiaoxia and Huang, Weilin and Huang, Yixuan and others},
  journal={arXiv preprint arXiv:2509.20427},
  year={2025}
}

@misc{Seedream4-5,
  title={Seedream 4.5},
  author={Seedream},
  howpublished={\url{https://seed.bytedance.com/en/seedream4_5/}},
  year={2025}
}

@misc{Seedream5-0,
  title={Seedream 5.0},
  author={Seedream},
  howpublished={\url{https://seed.bytedance.com/en/seedream5_0_lite/}},
  year={2026}
}

@misc{xia2025dreamomniunifiedimagegeneration,
      title={DreamOmni: Unified Image Generation and Editing}, 
      author={Bin Xia and Yuechen Zhang and Jingyao Li and Chengyao Wang and Yitong Wang and Xinglong Wu and Bei Yu and Jiaya Jia},
      year={2025},
      eprint={2412.17098},
      archivePrefix={arXiv},
      primaryClass={cs.CV},
      url={https://arxiv.org/abs/2412.17098}, 
}

@article{zhao2025envisioning,
  title={Envisioning beyond the pixels: Benchmarking reasoning-informed visual editing},
  author={Zhao, Xiangyu and Zhang, Peiyuan and Tang, Kexian and Zhu, Xiaorong and Li, Hao and Chai, Wenhao and Zhang, Zicheng and Xia, Renqiu and Zhai, Guangtao and Yan, Junchi and others},
  journal={arXiv preprint arXiv:2504.02826},
  year={2025}
}

@article{wang2025genexam,
  title={GenExam: A Multidisciplinary Text-to-Image Exam},
  author={Wang, Zhaokai and Yin, Penghao and Zhao, Xiangyu and Tian, Changyao and Qiao, Yu and Wang, Wenhai and Dai, Jifeng and Luo, Gen},
  journal={arXiv preprint arXiv:2509.14232},
  year={2025}
}

@article{Qwen3-VL,
      title={Qwen3-VL Technical Report}, 
      author={Shuai Bai and Yuxuan Cai and Ruizhe Chen and Keqin Chen and Xionghui Chen and Zesen Cheng and Lianghao Deng and Wei Ding and Chang Gao and Chunjiang Ge and Wenbin Ge and Zhifang Guo and Qidong Huang and Jie Huang and Fei Huang and Binyuan Hui and Shutong Jiang and Zhaohai Li and Mingsheng Li and Mei Li and Kaixin Li and Zicheng Lin and Junyang Lin and Xuejing Liu and Jiawei Liu and Chenglong Liu and Yang Liu and Dayiheng Liu and Shixuan Liu and Dunjie Lu and Ruilin Luo and Chenxu Lv and Rui Men and Lingchen Meng and Xuancheng Ren and Xingzhang Ren and Sibo Song and Yuchong Sun and Jun Tang and Jianhong Tu and Jianqiang Wan and Peng Wang and Pengfei Wang and Qiuyue Wang and Yuxuan Wang and Tianbao Xie and Yiheng Xu and Haiyang Xu and Jin Xu and Zhibo Yang and Mingkun Yang and Jianxin Yang and An Yang and Bowen Yu and Fei Zhang and Hang Zhang and Xi Zhang and Bo Zheng and Humen Zhong and Jingren Zhou and Fan Zhou and Jing Zhou and Yuanzhi Zhu and Ke Zhu},
	  journal={arXiv preprint arXiv:2511.21631},
      year={2025}
}

@misc{yu2022scalingautoregressivemodelscontentrich,
      title={Scaling Autoregressive Models for Content-Rich Text-to-Image Generation}, 
      author={Jiahui Yu and Yuanzhong Xu and Jing Yu Koh and Thang Luong and Gunjan Baid and Zirui Wang and Vijay Vasudevan and Alexander Ku and Yinfei Yang and Burcu Karagol Ayan and Ben Hutchinson and Wei Han and Zarana Parekh and Xin Li and Han Zhang and Jason Baldridge and Yonghui Wu},
      year={2022},
      eprint={2206.10789},
      archivePrefix={arXiv},
      primaryClass={cs.CV},
      url={https://arxiv.org/abs/2206.10789}, 
}

@misc{huang2023smarteditexploringcomplexinstructionbased,
      title={SmartEdit: Exploring Complex Instruction-based Image Editing with Multimodal Large Language Models}, 
      author={Yuzhou Huang and Liangbin Xie and Xintao Wang and Ziyang Yuan and Xiaodong Cun and Yixiao Ge and Jiantao Zhou and Chao Dong and Rui Huang and Ruimao Zhang and Ying Shan},
      year={2023},
      eprint={2312.06739},
      archivePrefix={arXiv},
      primaryClass={cs.CV},
      url={https://arxiv.org/abs/2312.06739}, 
}

@article{Show-o2,
  title={Show-o2: Improved Native Unified Multimodal Models},
  author={Xie, Jinheng and Yang, Zhenheng and Shou, Mike Zheng},
  journal={arXiv preprint arXiv:2506.15564},
  year={2025}
}

@misc{niu2025wiseworldknowledgeinformedsemantic,
      title={WISE: A World Knowledge-Informed Semantic Evaluation for Text-to-Image Generation}, 
      author={Yuwei Niu and Munan Ning and Mengren Zheng and Weiyang Jin and Bin Lin and Peng Jin and Jiaqi Liao and Chaoran Feng and Kunpeng Ning and Bin Zhu and Li Yuan},
      year={2025},
      eprint={2503.07265},
      archivePrefix={arXiv},
      primaryClass={cs.CV},
      url={https://arxiv.org/abs/2503.07265}, 
}

@misc{Gemini-3-Flash,
  title={Gemini 3 Flash: frontier intelligence built for speed},
  author={Google},
  howpublished={\url{https://blog.google/products-and-platforms/products/gemini/gemini-3-flash/}},
  year={2025}
}

@misc{Nano-Banana-Pro,
  title={Introducing Nano Banana Pro},
  author={Google},
  howpublished={\url{https://blog.google/innovation-and-ai/products/nano-banana-pro/}},
  year={2025}
}

@misc{Nano-Banana-2,
  title={Nano Banana 2: Combining Pro capabilities with lightning-fast speed},
  author={Google},
  howpublished={\url{https://blog.google/innovation-and-ai/technology/ai/nano-banana-2/}},
  year={2025}
}

@article{GPT-Image-1,
  title={GPT-Image-1},
  author={OpenAI},
  journal={https://openai.com/index/image-generation-api/},
  year={2025}
}

@article{GPT-Image-1.5,
  title={GPT-Image-1.5},
  author={OpenAI},
  journal={
https://openai.com/zh-Hans-CN/index/new-chatgpt-images-is-here/},
  year={2025}
}

@article{wang2025internsvg,
  title   = {InternSVG: Towards Unified SVG Tasks with Multimodal Large Language Models},
  author  = {Wang, Haomin and Yin, Jinhui and Wei, Qi and Zeng, Wenguang and Gu, Lixin and Ye, Shenglong and Gao, Zhangwei and Wang, Yaohui and Zhang, Yanting and Li, Yuanqi and Guo, Yanwen and Wang, Wenhai and Chen, Kai and Qiao, Yu and Zhang, Hongjie},
  journal = {arXiv preprint arXiv:2510.11341},
  year    = {2025},
  doi     = {10.48550/arXiv.2510.11341},
  url     = {https://arxiv.org/abs/2510.11341}
}

@article{sgpbench,
  title   = {Can Large Language Models Understand Symbolic Graphics Programs?},
  author  = {Qiu, Zeju and Liu, Weiyang and Feng, Haiwen and Liu, Zhen and Xiao, Tim Z. and Collins, Katherine M. and Tenenbaum, Joshua B. and Weller, Adrian and Black, Michael J. and Sch{\"o}lkopf, Bernhard},
  journal = {arXiv preprint arXiv:2408.08313},
  year    = {2024},
  url     = {https://arxiv.org/abs/2408.08313}
}

@article{paddleocr3,
  title   = {PaddleOCR 3.0 Technical Report},
  author  = {Cui, Cheng and Sun, Ting and Lin, Manhui and Gao, Tingquan and Zhang, Yubo and Liu, Jiaxuan and Wang, Xueqing and Zhang, Zelun and Zhou, Changda and Liu, Hongen and Zhang, Yue and Lv, Wenyu and Huang, Kui and Zhang, Yichao and Zhang, Jing and Zhang, Jun and Liu, Yi and Yu, Dianhai and Ma, Yanjun},
  journal = {arXiv preprint arXiv:2507.05595},
  year    = {2025},
  doi     = {10.48550/arXiv.2507.05595},
  url     = {https://arxiv.org/abs/2507.05595}
}

@article{irishman,
  title={Tunesformer: Forming irish tunes with control codes by bar patching},
  author={Wu, Shangda and Li, Xiaobing and Yu, Feng and Sun, Maosong},
  journal={arXiv preprint arXiv:2301.02884},
  year={2023}
}

@book{lakhmidi,
  title={Learning-based methods for comparing sequences, with applications to audio-to-midi alignment and matching},
  author={Raffel, Colin},
  year={2016},
  publisher={Columbia University}
}

@article{MAESTRO,
  title={Enabling factorized piano music modeling and generation with the MAESTRO dataset},
  author={Hawthorne, Curtis and Stasyuk, Andriy and Roberts, Adam and Simon, Ian and Huang, Cheng-Zhi Anna and Dieleman, Sander and Elsen, Erich and Engel, Jesse and Eck, Douglas},
  journal={arXiv preprint arXiv:1810.12247},
  year={2018}
}

@misc{qwen3max,
    title = {Qwen3-Max: Just Scale it},
    author = {Qwen Team},
    month = {September},
    year = {2025}
}

@misc{gptimage2,
  author       = {{OpenAI}},
  title        = {GPT-Image-2},
  year         = {2026},
  howpublished = {\url{https://openai.com/}},
  note         = {AI image generation model}
}

@article{lin2026scientific,
  title={Scientific Graphics Program Synthesis via Dual Self-Consistency Reinforcement Learning},
  author={Lin, Juekai and Zhu, Yun and Lin, Honglin and Li, Sijing and Lin, Tianwei and Liu, Zheng and Wang, Xiaoyang and Zhang, Wenqiao and Wu, Lijun},
  journal={arXiv preprint arXiv:2604.06079},
  year={2026}
}

@inproceedings{belouadi2024automatikz,
  title={Automatikz: Text-guided synthesis of scientific vector graphics with tikz},
  author={Belouadi, Jonas and Lauscher, Anne and Eger, Steffen},
  booktitle={International Conference on Learning Representations},
  volume={2024},
  pages={55917--55943},
  year={2024}
}

@misc{waltonfuture_k12,
  title        = {K12 Dataset},
  author       = {WaltonFuture},
  year         = {2026},
  howpublished = {\url{https://huggingface.co/datasets/WaltonFuture/K12}},
  note         = {Accessed: 2026-05-27}
}

@inproceedings{wang2025mv,
  title={Mv-math: Evaluating multimodal math reasoning in multi-visual contexts},
  author={Wang, Peijie and Li, Zhong-Zhi and Yin, Fei and Ran, Dekang and Liu, Cheng-Lin},
  booktitle={Proceedings of the Computer Vision and Pattern Recognition Conference},
  pages={19541--19551},
  year={2025}
}

@misc{zhang2024mavismathematicalvisualinstruction,
      title={MAVIS: Mathematical Visual Instruction Tuning}, 
      author={Renrui Zhang and Xinyu Wei and Dongzhi Jiang and Yichi Zhang and Ziyu Guo and Chengzhuo Tong and Jiaming Liu and Aojun Zhou and Bin Wei and Shanghang Zhang and Peng Gao and Hongsheng Li},
      year={2024},
      eprint={2407.08739},
      archivePrefix={arXiv},
      primaryClass={cs.CV},
      url={https://arxiv.org/abs/2407.08739}, 
}

@misc{
huang2026chemeval,
title={ChemEval: A Multi-level and Fine-grained Chemical Capability Evaluation for Large Language Models},
author={Yuqing Huang and Rongyang Zhang and Xuesong He and Xuyang Zhi and Hao Wang and Nuo Chen and Liuzongbo and Xin Li and Feiyang Xu and Deguang Liu and Huadong Liang and YiLi and Jian Cui and Yin Xu and Shijin Wang and Guiquan Liu and Qi Liu and Defu Lian and Enhong Chen},
year={2026},
url={https://openreview.net/forum?id=dVb96Endxa}
}

@book{ahern2018biochemistry,
  author    = {Ahern, Kevin and Rajagopal, Indira and Tan, Taralyn},
  title     = {Biochemistry Free For All},
  publisher = {Oregon State University},
  address   = {Corvallis, OR},
  year      = {2018},
  url       = {https://open.oregonstate.education/biochemfreeforall/},
  urldate   = {2026-06-22},
  note      = {Open textbook licensed under a Creative Commons Attribution-NonCommercial license}
}

@book{keenleyside2019microbiology,
  author    = {Keenleyside, Wendy},
  title     = {Microbiology: Canadian Edition},
  publisher = {eCampusOntario},
  address   = {Toronto, ON},
  year      = {2019},
  url       = {https://ecampusontario.pressbooks.pub/microbio/},
  urldate   = {2026-06-22},
  note      = {First edition; derived from OpenStax Microbiology; licensed under CC BY 4.0, except where otherwise noted}
}

@misc{kimball2025biology,
  author       = {Kimball, John W.},
  title        = {Kimball's Biology Pages},
  year         = {2025},
  howpublished = {\url{https://www.biology-pages.info/}},
  urldate      = {2026-06-22},
  note         = {Online biology textbook/reference distributed under CC BY 3.0}
}

@book{openstax_micro_2e,
  author    = {Steven A. Greenlaw and David Shapiro},
  title     = {Principles of Microeconomics 2e},
  year      = {2017},
  publisher = {OpenStax},
  address   = {Houston, TX},
  url       = {https://openstax.org/books/principles-microeconomics-2e/pages/1-introduction},
  note      = {Open textbook, licensed under CC BY 4.0}
}

@book{openstax_macro_2e,
  author    = {Steven A. Greenlaw and David Shapiro},
  title     = {Principles of Macroeconomics 2e},
  year      = {2017},
  publisher = {OpenStax},
  address   = {Houston, TX},
  url       = {https://openstax.org/books/principles-macroeconomics-2e/pages/1-introduction},
  note      = {Open textbook, licensed under CC BY 4.0}
}

@book{openstax_econ_2e,
  author    = {Steven A. Greenlaw and David Shapiro},
  title     = {Principles of Economics 2e},
  year      = {2017},
  publisher = {OpenStax},
  address   = {Houston, TX},
  url       = {https://openstax.org/books/principles-economics-2e/pages/1-introduction},
  note      = {Open textbook, licensed under CC BY 4.0}
}

\appendix
\clearpage

\section{Dataset Construction Details}
\label{sec:appendix}

\subsection{Sub-Discipline Distribution}

\begin{figure}[h]
  \centering
  \includegraphics[width=0.75\linewidth]{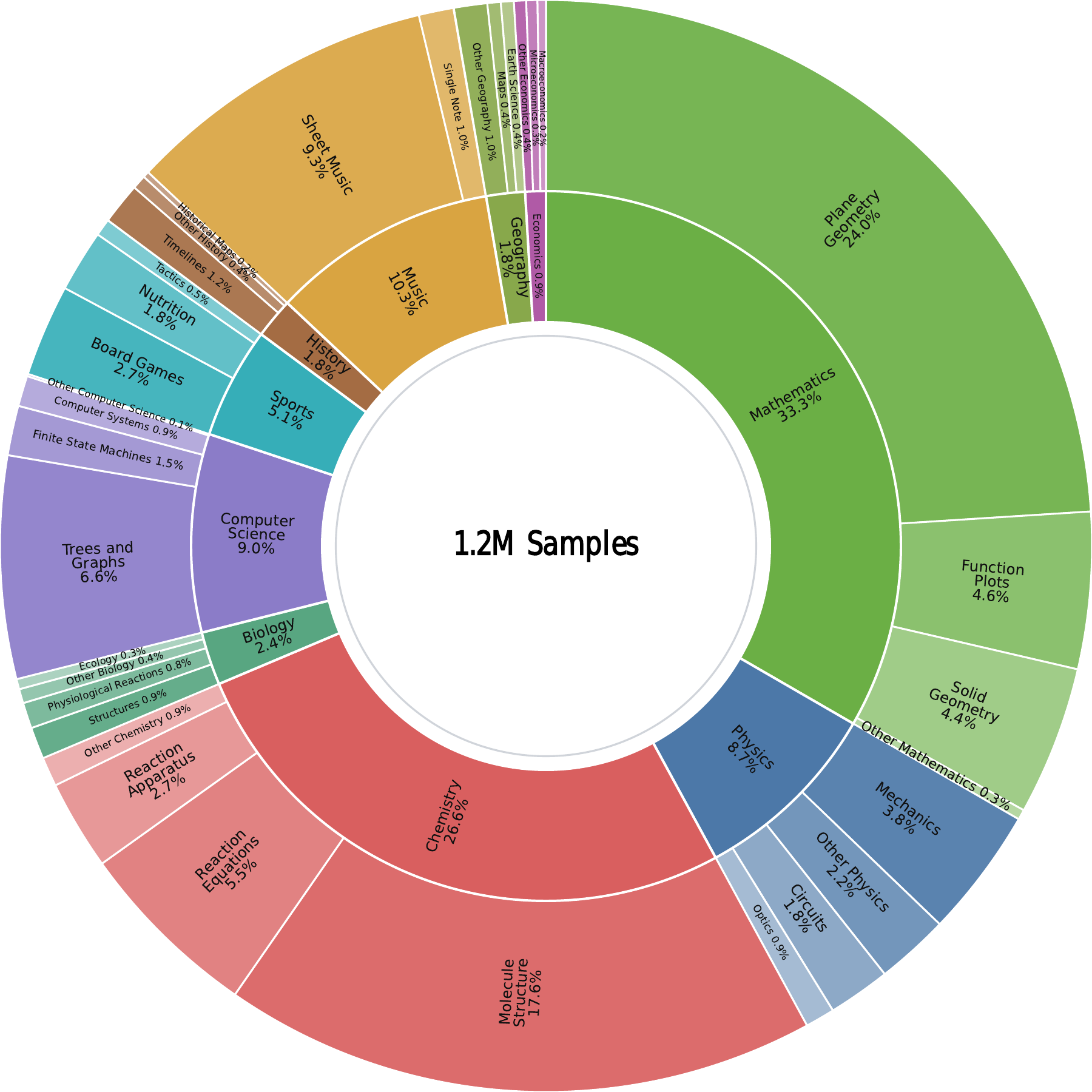}
  \caption{Sub-discipline distribution of \datasetname.}
  \label{fig:subdiscipline}
\end{figure}

Figure~\ref{fig:subdiscipline} shows the sub-discipline composition of \datasetname, covering 1.2M samples in total. The distribution is intentionally broad but not uniform. Mathematics and chemistry form the two largest discipline groups, accounting for 33.3\% and 26.6\% of the data, respectively, because these domains provide many programmatically verifiable structures such as plane geometry, function plots, solid geometry, molecular structures, and reaction equations. Music, computer science, and physics contribute another substantial portion through sheet music, trees and graphs, finite-state machines, mechanics, circuits, and optics. The long tail covers sports, biology, history, geography, and economics, which increases visual and semantic diversity beyond the dominant scientific diagram types. At the sub-discipline level, large categories such as plane geometry, molecular structures, and sheet music are complemented by smaller but distinctive formats including maps, timelines, nutrition charts, finite-state machines, ecology diagrams, and macroeconomics plots. This mixture supports both scale and breadth: high-frequency categories provide enough examples for robust structural learning, while low-frequency categories expose models to specialized academic visual conventions.

\subsection{Specialized Synthesis Pipelines}

\subsubsection{Mathematics Data Synthesis}
We organize the mathematics data construction into three major categories: functions, plane geometry, and solid geometry.

\textbf{(1) Functions.} Data are synthesized using the Matplotlib library to render a diverse set of function plots. The pipeline covers five task types: \textbf{\romannumeral 1) Symmetry}, where plots of functions are displayed under odd/even transformations; \textbf{\romannumeral 2) Zeros and Poles}, where the function's roots and poles are highlighted; \textbf{\romannumeral 3) Tangent Lines}, where the tangent at a specified point is drawn; \textbf{\romannumeral 4) Derivatives}, where the plot of the derivative is rendered alongside the original function; \textbf{\romannumeral 5) Function Properties}, where parameters of trigonometric functions such as amplitude, period, and phase are varied to generate diverse behaviors.

\textbf{(2) Plane Geometry.} Plane geometry data are generated and rendered using Matplotlib and MathCanvas~\cite{shi2025mathcanvasintrinsicvisualchainofthought}, producing diverse planar figures for downstream editing tasks. The pipeline covers three task types: \textbf{\romannumeral 1) Points and Lines}, where constructions such as perpendiculars or parallels through a point outside a line are drawn; \textbf{\romannumeral 2) Geometric Transformations}, including rotation, translation, and reflection of planar figures; and \textbf{\romannumeral 3) Geometric Properties}, such as medians and altitudes of triangles, tangents to circles, and angle bisectors within plane shapes.

\textbf{(3) Solid Geometry.} We generate solid geometry data using GeoGebra~\cite{geogebra:website} and Matplotlib to construct a variety of 3D editing tasks. The pipeline covers four types of operations: \textbf{\romannumeral 1) Solid Symmetry}, where a solid is mirrored either across a plane or with respect to a point; \textbf{\romannumeral 2) Solids of Revolution}, where a shape is rotated around an axis; \textbf{\romannumeral 3) Solid Translation}, where a solid is shifted along a vector; and \textbf{\romannumeral 4) Solid Projection}, where the orthogonal projection of a solid onto the $x$-$y$ plane is rendered.

\subsubsection{History Data Synthesis}
We construct two types of editing tasks for the history domain.

\textbf{(1) Timeline Completion.} We collect historical event entries from Wikimedia Commons and retrieve their associated images through the static resource. To ensure that each sample is grounded in structured historical metadata, we only retain event entries that are linked to at least one Wikimedia Commons image. The resulting event pool contains approximately 500 historical events spanning from 3000 BCE to 1900 CE.

During data generation, we randomly sample a batch of events from the pool and sort them chronologically to construct a timeline. The timeline is rendered with Matplotlib, where visual attributes such as colors, layouts, fonts, and line styles are randomly configured to improve visual diversity. For each rendered timeline, one event is randomly masked to produce the input image, while the complete timeline is used as the ground-truth image. This process yields 10K paired editing samples for the timeline completion task.

\textbf{(2) Historical Map Completion.} The input and ground-truth map images are rendered using a Python-based geospatial visualization pipeline, primarily built upon Matplotlib and GeoPandas for rasterizing GeoJSON vector data. To improve visual diversity, we adopt multiple theme-specific color palettes and randomly sample rendering styles, producing maps with varied background colors, region colors, label styles, and visual layouts. Given the input GeoJSON data, the pipeline first performs random spatial cropping to generate local map regions. To control the difficulty of each sample, we further apply an adaptive scale search strategy, where the cropping ratio is iteratively adjusted through binary search until the number of visible labels falls within a predefined range, i.e., 5-8 labels. This ensures that each generated map contains sufficient textual content while avoiding overly crowded layouts.

To improve label readability, we introduce an iterative label de-overlapping algorithm. Specifically, the pipeline computes the spatial overlap between text bounding boxes and applies repulsive offsets to conflicting labels, enabling automatic label avoidance and more balanced text placement. The final ground-truth image preserves all visible labels, while the input image masks a subset of labels to construct paired samples for historical map completion.

\subsubsection{Music Data Synthesis}

We construct image editing and text-to-image tasks for the music domain, covering sheet music, piano roll, and music theory chart representations.

\textbf{(1) Single-Note Editing and Generation.} We programmatically generate single-note data, acting as a simple level, to strengthen perception of fundamental symbols. Editing tasks cover transposition, duration change, accidental addition, and dotting; a notehead-only variant hides the staff to focus on duration symbol transformations. A complementary text-to-image task covers the full combination of pitches, clefs, and durations.

\textbf{(2) Text-to-Score Generation.} This pipeline adopts a three-tier strategy: exhaustive enumeration of fundamental elements (key/time signatures, scales, chords, intervals, etc.), random sampling of valid multi-measure passages with diatonic and stepwise motion constraints, and reverse-engineered note-level descriptions from IrishMAN ABC tunes~\cite{irishman}. 

\textbf{(3) Sheet Music Editing.} ABC notation folk tunes from IrishMAN are parsed and segmented into measure-level chunks. Deterministic editing operations---transposition, clef change, dynamic marking and ornament insertion---are applied to each chunk and rendered via LilyPond, with pre- and post-edit images forming paired samples.

\textbf{(4) Piano Roll Editing.} MIDI files from Lakh MIDI~\cite{lakhmidi} and MAESTRO~\cite{MAESTRO} are parsed and edited at the note level (transposition, time stretching, rhythmic quantization), then rendered as DAW-style piano roll images. A matrix-level difference check ensures each operation produces visible changes.

\textbf{(5) Music Theory Chart Editing.} Piano keyboard diagrams are programmatically generated, highlighting chord or scale notes on a two-octave keyboard. Editing operations include transposition, chord quality transformation, and mode change across all seven church modes.

\textbf{(6) Cross-Representation Conversion.} Each IrishMAN ABC score segment is simultaneously rendered as both sheet music and piano roll, forming cross-representation conversion pairs.

\subsubsection{Computer Science Data Synthesis}
  We build a data synthesis engine for the computer science domain using the Python libraries, covering 17 editing and reasoning tasks across three fundamental data structures: trees, graphs, and finite state machines (FSMs). Specifically, the tasks include:

    \textbf{(1) Tree:} Topology editing and node manipulation, traversal path visualization, binary search tree (BST) operations, heap operations with dual tree/array views, Huffman coding tree construction, and lowest common ancestor (LCA) finding and path highlighting.
   
   \textbf{(2) Graph:} K-hop neighborhood identification, node degree identification, cycle detection, bipartite graph coloring, shortest path reasoning, and directed graph reachability analysis.
      
      \textbf{(3) FSM:} Input string tracing, state role identification, and transition logic completion.

Rendering engines are selected according to task characteristics, with Matplotlib used for trees and graphs and Graphviz (\texttt{neato}) for FSMs. To preserve spatial consistency between edited image pairs, node layouts are fixed before modification. We further apply automatic quality control to filter invalid or overlapping samples. Chain-of-thought (CoT) annotations are generated synchronously by deterministic solvers and later refined by a large language model (Qwen-Max~\cite{qwen3max}).

The computer science domain covers 17 visual editing and reasoning tasks organized into three categories: trees, graphs, and finite state machines. All samples are programmatically generated via rule-based engines without reliance on external datasets. Each task supports both the img2img (paired editing images) and text2img (text description to target image) paradigms, producing approximately 64,000 samples per paradigm and roughly 128,000 in total.

Rendering backends are selected based on structural characteristics. Matplotlib is used for tree and general graph tasks because it supports efficient large-scale batch rendering and flexible programmatic control. For finite-state machines (FSMs), we adopt Graphviz with the \texttt{neato} engine to obtain clearer layouts in dense state-transition diagrams.

To maintain spatial consistency within each edited image pair, node coordinates are anchored prior to applying any modification, ensuring that unchanged components preserve identical positions across the before and after images.

We additionally incorporate a geometric collision detection module into the generation pipeline. Node-to-node and node-to-edge distances are explicitly computed, and samples violating predefined minimum-clearance constraints are automatically discarded and regenerated. This procedure substantially improves overall visual quality and reduces overlapping artifacts.

Chain-of-thought (CoT) annotations are produced synchronously using deterministic solvers during data generation and are subsequently rewritten by a large language model (Qwen-Max~\cite{qwen3max}) to improve readability and linguistic diversity.

\textbf{(1) Tree Topology Editing.} We programmatically generate full binary trees and random general trees, then apply two topology-level operations: completing an incomplete binary tree to a full binary tree of a specified depth, and pruning the subtree rooted at a designated node. Before and after images share the same layout anchors so that unmodified regions remain visually consistent. This task shares a generator with the next one, producing approximately 6,000 pairs combined.

\textbf{(2) Tree Node Manipulation.} Given a randomly generated tree, a parent node satisfying branching-factor constraints is selected and a new child node is inserted with a specified color. The instruction explicitly states the target node ID, parent node, directional placement (left/right child), and node color.

\textbf{(3) Traversal Path Visualization.} Random binary trees are generated and traversed in pre-order, in-order, or post-order. The ground-truth image randomly adopts one of two visualization styles: a smooth spline curve connecting red marker dots along the traversal path, or red circular badges numbering the visit order beneath each node. Approximately 6,000 pairs.

\textbf{(4) Binary Search Tree Operations.} Random BSTs are constructed and subjected to two types of operations: inserting new values with the newly added nodes highlighted by red rectangular boxes, or swapping two node values to violate the BST property, requiring identification and correction of the erroneous positions. The insert-to-fix ratio is approximately 7:3, yielding about 6,000 pairs.

\textbf{(5) Heap Operations.} Random max-heaps or min-heaps are generated, and a single insertion or root extraction is performed. The ground-truth image presents a dual view: a tree view of the heap alongside a bottom array view, with tree nodes and array cells linked by a consistent color mapping. Approximately 4,000 pairs.

\textbf{(6) Huffman Coding Tree Construction.} A set of characters with random frequency weights is generated. The input image displays the scattered leaf nodes, while the ground-truth image shows the complete Huffman tree built by greedy merging, annotated with cumulative weights at internal nodes and 0/1 coding labels on edges. Approximately 2,000 pairs.

\textbf{(7) Lowest Common Ancestor Finding.} Random trees (binary or ternary) are generated, and two descendant nodes from distinct branches are selected. The ground-truth image marks their LCA with a red rectangular box. This task shares a generator with the next one, producing approximately 4,000 pairs combined.

\textbf{(8) Tree Path Highlighting.} On the same type of random tree, two leaf nodes are selected and the ground-truth image highlights the shortest path between them with a thickened red line.

\textbf{(9) K-Hop Neighborhood Identification.} Random connected undirected graphs are generated. Given a start node and distance $k$, the ground-truth image marks the start node with a gold border and all nodes exactly $k$ hops away with green borders. Approximately 4,000 pairs.

\textbf{(10) Node Degree Identification.} Given a target degree on a random undirected graph, the ground-truth image highlights all qualifying nodes with green rectangular boxes. Approximately 4,000 pairs.

\textbf{(11) Cycle Detection.} A graph containing exactly one cycle is constructed by adding a single extra edge to a random spanning tree. The ground-truth image highlights all nodes and edges on the cycle in red. Approximately 4,000 pairs.

\textbf{(12) Bipartite Graph Coloring.} Random bipartite graphs are generated with all nodes rendered in grey in the input image. The ground-truth image assigns two distinct colors to the two independent sets such that no adjacent nodes share the same color. Approximately 4,000 pairs.

\textbf{(13) Shortest Path Reasoning.} Given a start and end node on a random connected undirected graph, the ground-truth image highlights all nodes and edges along the BFS shortest path. Approximately 4,000 pairs.

\textbf{(14) Directed Graph Reachability Analysis.} Random directed graphs are generated. Given a start node, the ground-truth image marks all nodes reachable via directed edges with red dashed rectangular borders. Approximately 4,000 pairs.

\textbf{(15) DFA Input String Tracing.} A random deterministic finite automaton (DFA) is generated along with an input string. The ground-truth image highlights the traversed transition edges in blue and fills the final state with green (accepted) or red (rejected). Approximately 4,000 pairs.

\textbf{(16) State Role Identification.} On a randomly generated FSM where all states are initially rendered in grey, the ground-truth image assigns distinct colors to the start state, accept states, and regular states respectively. Approximately 4,000 pairs.

\textbf{(17) Transition Logic Completion.} One transition edge is hidden from a complete DFA. The input image shows the incomplete automaton, and the ground-truth image highlights the restored missing transition in red. Approximately 4,000 pairs.

Tasks 1--14 use NetworkX for topology construction and Matplotlib for rendering; tasks 15--17 use the Graphviz \texttt{neato} engine. All images are rendered with randomized combinations of 14 color themes and multiple font families to increase visual diversity.

\subsubsection{Chemistry Data Synthesis}
We construct three types of editing tasks for the chemistry domain.

\textbf{(1) Molecule Generation.} This part of the data is extracted from the ChEMBL database~\cite{10.1093/nar/gkad1004}. Text-to-image data are generated from the SMILES~\cite{Weininger1988} formulas of substances using RDKit~\cite{rdkit}. We adopt diverse template descriptions together with the given substance name as the prompt, and use the corresponding generated 2D molecular structure image as the ground truth.

\textbf{(2) Reaction Equation Editing.} The data is extracted from USPTO-50K~\cite{Schneider2016} and common chemical reactions from the Internet. Two rendering engines, RDKit~\cite{rdkit} and Indigo~\cite{Pavlov2011}, and seven rendering styles are used to enhance data diversity. The pipeline covers two sub-tasks, both of which use the reactant molecular structures as the input image, while the ground truth is divided into two types: the complete reaction equation and the major product molecule.

\textbf{(3) Molecular Structure Editing.} Molecular structure editing data is constructed based on the ChEMBL database~\cite{10.1093/nar/gkad1004} by inserting functional-group fragments into the original molecules to generate editing tasks. The pipeline includes various groups such as phenyl, methoxy, and methylamino, together with insertion validity checking. It supports both single-site editing and multi-site editing with different corresponding positions in the original molecule highlighted in different colors.

\subsubsection{Sports Data Synthesis}

We construct the sports portion of the dataset from four subdomains: Chess/Xiangqi, Go, sports nutrition, and soccer tactics. \footnote{
  Lichess: \url{https://database.lichess.org/standard/};
  XQBase: \url{https://www.xqbase.com/xqbase/};
  KataGo Archive: \url{https://katagoarchive.org/kata1};
  USDA FoodData Central: \url{https://fdc.nal.usda.gov/};
  GI labels: \url{https://www.gluok.com/};
  StatsBomb Open Data: \url{https://github.com/statsbomb/open-data}.
  }

\textbf{(1) Chess/Xiangqi:} We collect game records from Lichess and XQBase, parse them into structured move sequences, and generate \textit{opening},
\textit{legal\_moves}, and \textit{bestmove} editing tasks, where the last task is further filtered by engine-based verification using Stockfish
or Pikafish. 

Specifically, we construct chess and xiangqi image-editing data from game records. For chess, the raw source is the Lichess standard-game PGN archive. For
xiangqi, the raw source is XQBase game pages, which are crawled and parsed into structured JSONL records containing opening names, move sequences,
and source identifiers. Based on these structured records, we build three task types: \textit{opening}, \textit{legal\_moves}, and
\textit{bestmove}. The \textit{opening} task maps early-game positions to canonical opening states. The \textit{legal\_moves} task highlights all
legal destinations for a selected piece. The \textit{bestmove} task uses external engines, i.e., Stockfish for chess and Pikafish for xiangqi, and
keeps only positions whose best moves remain stable under deeper verification and are sufficiently separated from runner-up moves. All samples are
rendered into paired \texttt{editing} and \texttt{gt} images, together with a structured JSON file.

\textbf{(2) Go:} We extract candidate board states from KataGo-related SGF archives and further relabel them with KataGo analysis to obtain
reliable \textit{crucial\_move} supervision. 

Specifically, we construct Go image-editing data from KataGo-related SGF archives. We first parse SGF game trees and sample candidate positions within
controlled ply ranges, using the next SGF move as a weak label. We then relabel these candidates with local KataGo analysis to obtain stronger
supervision. Positions are retained only when the engine-selected top move is sufficiently reliable, according to thresholds on visits, score gap,
and win-rate gap, and when the position is not already too one-sided. The final Go task is \textit{crucial\_move}, where the input is the current
board state and the target image adds the key next move marked with ``1''.

\textbf{(3) Sports Nutrition:} We build a unified food asset table from USDA FoodData Central together with manually curated GI labels and food images, and render multiple nutrition-oriented editing tasks such as grouping, pyramid construction, pie-chart
integration, and attribute highlighting. 

Specifically, the sports nutrition pipeline is built from USDA FoodData Central, combined with manually curated food lists, glycemic-index references, and local
food images or transparent cutouts. These resources are merged into a unified asset table that stores nutritional attributes, macro categories, GI
annotations, and image paths. Based on this asset table, we generate several synthetic image-editing tasks, including \textit{classify\_grouping},
\textit{pie\_chart\_integration}, \textit{nutrition\_pyramid}, \textit{highlight\_high\_gi}, and \textit{highlight\_high\_protein}. The resulting
images are fully programmatically rendered while remaining grounded in real nutritional data.

\textbf{(4) Soccer Tactics:} We parse StatsBomb Open Data into canonical tactical records and render formation- and ball-handler-centered editing tasks on synthetic tactics boards. All generated samples are finally converted into a unified multimodal editing
format with paired input and target images, textual editing instructions, and structured metadata.

Specifically, the soccer tactics pipeline mainly uses StatsBomb Open Data, with optional auxiliary support from Metrica sample tracking data. We first parse
match events and tactical information into a canonical JSONL format containing formations, player coordinates, ball-handler identity, and other
metadata. Based on these structured records, we generate several tactics-board editing tasks, including \textit{soccer\_formation\_dots},
\textit{soccer\_formation\_jerseys}, and \textit{soccer\_ball\_handler\_highlight}. These tasks do not require external search engines; the
supervision is directly derived from structured tactical annotations.

\textbf{Unified Annotation Format.}
Although the raw sources and rendering processes differ across sports subdomains, all generated samples are finally converted into a unified
multimodal editing format. Each record contains a conditioning image, a target image, a natural-language editing instruction, image size metadata,
generation flags, and an \texttt{original} field storing task-specific metadata. Large datasets are automatically sharded into multiple JSONL
files, and a corresponding \texttt{meta.json} file records dataset-level information such as the root directory, annotation path, number of
shards, and task type.

\textbf{Instruction Diversification.}
We further support prompt diversification by rewriting a subset of instructions with semantically equivalent templates while keeping the image
supervision unchanged. This increases linguistic variety without altering the underlying visual-editing objective.

\subsection{SVG-Based Data Rendering}

\textbf{Rendering from Structured Semantic Instance.} SVG-Based Data Rendering provides a structured-knowledge-driven pipeline for constructing discipline-specific image generation and image editing data in economics and biology.
As input sources, representative knowledge points and typical diagrammatic content are first manually selected from openly licensed textbooks and educational resources~\citep{ahern2018biochemistry,keenleyside2019microbiology,kimball2025biology,openstax_micro_2e,openstax_macro_2e,openstax_econ_2e}. In the transformation process, GPT-5~\citep{singh2025openai} is used as a structuring assistant to convert the selected content into machine-readable knowledge units, including fields such as topic, objects, textual elements, relational information, diagram type, coordinate semantics, and visual constraints. The system then invokes predefined SVG templates, parameter samplers, and rendering rules to automatically generate disciplinary images such as tables, flowcharts, classification trees, quadrant diagrams, and coordinate-curve plots, together with corresponding captions. For quality control, the structured knowledge units and rendered images are manually checked to ensure disciplinary correctness, visual clarity, and consistency with the source knowledge. The final outputs include image-caption pairs for image generation tasks and triplet samples for image editing tasks, consisting of the original image, editing instruction, and edited image. This process decouples disciplinary knowledge modeling from visual rendering, improving the controllability, interpretability, and subject consistency of the constructed data.

The SVG-based framework is instantiated for six diagram families: biological lifecycle diagrams, economics curve charts, biological curve charts, biochemical reaction equations, food web diagrams, and food chain diagrams. For lifecycle, food chain, and food web diagrams, the main visual elements are English text nodes connected by arrows or relations, rather than natural-image objects. For reaction equations, editable elements are textual reactants or products. For curve charts, editable elements include curve parameters, axes, key points, annotations, and domain-specific semantic states. Therefore, all diagrams are treated as structured scientific visualizations composed of graphical primitives and textual elements.

\textbf{Rendering from Web Images.}
For web images, we use PaddleOCR~\cite{paddleocr3} to extract images and their associated textual context. We then apply Qwen3-VL-8B~\cite{Qwen3-VL} to remove samples from undesirable domains, such as illustrative textbook figures, real-world photographs in exam questions, mathematical formulas, and visually complex images that are difficult to convert into SVG format.

Next, each image and its context are provided to Gemini-3-Flash~\cite{Gemini-3-Flash} to produce the SVG code of the original image. The model further determines whether the original image should be treated as the input or the output image, and generates both an editing prompt and the SVG code of the paired image. The two SVG files are then rendered into an image pair. The prompts generated by Gemini-3-Flash~\cite{Gemini-3-Flash} are detailed and explicit, and can serve as step-by-step reasoning prompts. To evaluate discipline-specific knowledge and competence, we further extract and summarize the key information in these prompts to obtain simplified, implicit prompts as the final data.

Finally, we perform an additional filtering step on each triplet of <input\_image, prompt, output\_image> to assess data correctness and quality. Gemini-3-Flash~\cite{Gemini-3-Flash} is used to remove samples containing rich text formatting, physical structural errors, or annotation errors.

This construction strategy provides several practical advantages. First, it reduces the cost of data collection and annotation by automatically generating images, captions, metadata, and editing instructions from reusable structured representations. Second, it improves controllability since both visual appearance and semantic content can be explicitly specified and sampled. Third, it enables fine-grained and verifiable editing, as each edit corresponds to a structured transformation rather than an uncontrolled pixel-level modification. Finally, it supports scalable dataset expansion: new diagram types, templates, scientific constraints, or editing operations can be incorporated without redesigning the entire data generation pipeline.

\subsection{TikZ-Based Data Rendering}

\subsubsection{Method}

We construct text-to-image generation and image editing tasks across
Biology, Geography, Physics, Chemistry, Computer Science, and
Economics, with visual representations spanning anatomical and
structural diagrams, flowcharts, plotted curves, network and pedigree
figures, geological cross-sections, directional maps, and ray-optics
/ circuit schematics. All pipelines draw from a shared LLM-generated
seed library of 1346 items covering 17 fine-grained
knowledge points across the six disciplines (e.g.\ cell-structure
``fill-in'', ``ray optics'', ``supply-and-demand curves''); each item
carries common fields (discipline, category, difficulty, knowledge-point
name) plus per-KP custom fields that are flattened into a single
topic string for downstream rendering.

\textbf{(1) TikZ-Compiled Generation.} Each seed is rewritten by the
LLM into a drawing blueprint with discipline-specific LaTeX package
hints, then translated into TikZ source. Inspired by the tool-use
formulation of~\cite{lin2026scientific}, \texttt{pdflatex} is exposed
as an external tool the model iterates against: compile errors are
fed back to repair the source (inner self-repair loop). Successful
renders are scored by a vision-model judge along correctness,
clarity, and completeness; failures send feedback and the previous
PNG back to code generation (outer visual-retry loop). The
description and final PNG form a text-to-image pair.

\textbf{(2) TikZ Label Fill-in Editing.} For seeds
whose main diagram has passed scoring and carries a labels list
(e.g.\ component names on a cell-structure figure), we apply a
deterministic regex rewrite to the final \texttt{.tex}---replacing
each label in place with numbered placeholders in first-occurrence order---and recompile to
obtain a geometrically aligned blanked PNG and an answer key. Because
both diagrams share one \texttt{.tex} source, they are
pixel-identical except for the annotation tokens, forming a natural
image-editing pair usable in both directions (annotated
$\leftrightarrow$ blanked) without alignment checks.

  \textbf{Seed Library Construction.} The 17 knowledge-point expansion
  prompts are hand-authored once per KP; the LLM is then driven through
   multi-pass continuation under a per-pass batch budget to bypass
  gateway-side truncation on long responses, with names already
  produced relayed back so the model proposes only new items, and
  deduplication performed on the merged result. The resulting per-item
  record carries common fields and a handful of KP-specific custom
  fields such as \texttt{labels}, \texttt{key\_points}, \texttt{view},
  and \texttt{curve\_type}, which the downstream loader concatenates
  with the item name into a single topic string.

  \textbf{Two-Track Prompt Families.} The TikZ and image-model
  pipelines maintain strictly disjoint per-discipline prompt packages: the
   TikZ side ships LaTeX macro-package hints (e.g.\
  \texttt{circuitikz}, \texttt{chemfig}, \texttt{chronosys}) while the
  image side ships pure art-direction guidance with all LaTeX and
  coordinate vocabulary explicitly forbidden (e.g.\ for Biology:
  ``mitochondria: bean-shaped/ellipsoidal, double membrane, cristae
  clearly visible''). Cross-contamination at the description stage is
  avoided by template-level isolation.

  \textbf{Reflection-Style Feedback Loops.} We observe that one-shot
  generations rarely pass quality control in either pipeline, but
  explicitly relaying the compiler's exact errors or the judge's
  pinpointed issues and corrective suggestions back to the generator
  substantially improves subsequent attempts, consistent with iterative
  self-refinement patterns observed in recent LLM systems.
  Both the inner compile-repair loop in (1) and the outer visual-retry
  loop shared by (1) and (3) are instances of this pattern.

  \textbf{Vision-Judge Rubric.} The judge returns JSON containing three
   sub-scores (correctness, clarity, completeness) and an aggregate,
  plus two mandatory list fields \texttt{knowledge\_errors} and
  \texttt{overlap\_issues} that act as hard vetoes: a non-empty entry in
   either fails the diagram regardless of the aggregate. The
  regeneration feedback prepends veto items ahead of the free-form
  summary string to guarantee they cannot be silently dropped by the
  judge's natural-language paraphrase.

  \textbf{Retries and Parallelism.} The TikZ pipeline uses nested
  retries (inner compile/repair $\times$ outer visual regeneration);
  the image pipeline retains only the outer loop. Both limits are
  configurable. Batch execution uses a thread-pool across topics, with
  per-topic exceptions absorbed into failure records so a single bad
  topic does not abort the run. Upstream seed expansion is constrained
  by API-gateway truncation on long responses and concurrency
  throttling, which forced a fallback to two parallel workers and
  multi-pass continuation with a per-response batch-size hint.

  \textbf{Trace and Reproducibility.} Every LLM prompt, raw response,
  extracted TikZ source, compile log, and intermediate PDF/PNG is
  written to disk under a monotonic-counter filename encoding step,
  visual round, and compile attempt; a chronological directory listing
  replays the full pipeline run. A per-topic JSON metadata record
  stores topic, description, scores, retry counts, failure reason, and
  fill-in outcome, allowing successful topics to be skipped on resume
  while partial runs are wiped clean to keep the trace counter
  monotonic.

  \textbf{Dataset Export.} Successful topics are aggregated into a
  JSONL dataset with relative image paths and automatic sharding once a
   single shard would exceed 50{,}000 lines. The natural-language
  description from step 1 is written into a chain-of-thought field for
  downstream training, and fill-in items additionally ship their answer
   keys. PNG dimensions are read directly from the IHDR chunk so the
  entire stack remains free of third-party Python dependencies.

\subsubsection{Prompt}
\label{appendix:tikz-prompts}

This section reproduces the full prompts for the LLM stages of the
TikZ-compiled generation track (track~(1) in the Method). That track drives
the LLM through five staged prompts: a description-expansion stage, a
code-generation stage, a compile-repair stage, and a two-step visual-quality
loop (a vision judge plus a feedback-driven regeneration). The stage numbers
below follow the pipeline's internal labels; Stage~3 is the \texttt{pdflatex}
compilation step itself, which issues no LLM prompt and is therefore omitted
here. All prompts are issued in English. Braced tokens such as
\texttt{\{topic\}}, \texttt{\{description\}}, \texttt{\{error\_log\}},
\texttt{\{code\}}, \texttt{\{prev\_code\}}, \texttt{\{feedback\}} and
\texttt{\{threshold\}} are filled at run time; \texttt{\{subject\_hints\}} is
replaced by the per-discipline package list reproduced at the end of this
section.

\paragraph{Stage 1 --- Description Expansion.}
Turns a knowledge point into a concrete, TikZ-shaped drawing description.

\begin{promptlisting}
You are an expert in {subject} education-diagram design.

Expand the following knowledge point into a **concrete TikZ drawing description**.

Requirements:
1. Describe which geometric primitives to draw (rectangles, circles, arrows, curves, filled regions, ...)
2. Specify relative positions and size relationships among the elements
3. List every English text/symbol annotation
4. If color coding is needed, state the palette
5. State the overall layout (landscape/portrait, approximate dimensions)

{subject_hints}

Knowledge point: {topic}

Output exactly one paragraph of drawing description. Do not output code. No extra content.
\end{promptlisting}

\paragraph{Stage 2 --- TikZ Code Generation.}
Turns the drawing description into a complete, compilable standalone document.

\begin{promptlisting}
You are a professional LaTeX/TikZ scientific-diagram developer. Generate TikZ code from the following drawing description.

Strict requirements:
1. Must be a complete, directly compilable standalone document
2. Begin with \documentclass[border=10pt]{standalone}
3. Declare every required package in the preamble
4. Forbidden: \includegraphics, \input or any external dependency
5. Target compiler is pdflatex
6. All in-figure text must be in English
7. Output only the code (inside a code block); no explanation

{subject_hints}

General guidance:
- Do NOT nest a circuitikz environment inside a tikzpicture — they are siblings
- Verify all coordinate computations to avoid element overlap
- Declare \usetikzlibrary{arrows.meta,patterns,positioning} when needed

Drawing description:
{description}
\end{promptlisting}

\paragraph{Stage 4 --- Compile Repair.}
Fixes code that failed to compile, from a filtered \texttt{pdflatex} error log.
(Retried up to a configurable number of times before the topic is abandoned.)

\begin{promptlisting}
The following TikZ code failed to compile. Fix it from the error log.

Requirements:
1. Make only the minimal necessary change — do not rewrite the whole file
2. Output the complete fixed code (not a patch)
3. Output only the code (inside a code block); no explanation
4. Target compiler is pdflatex
5. If the error stems from using a specialized component (circuitikz's ammeter/resistor, chemfig's molecular commands, chronosys's time events, etc.) inside the wrong environment or with a missing package — switch to the correct specialized environment / load the right package, do NOT delete the component

{subject_hints}

Error log (key lines):
```
{error_log}
```

Original code:
```latex
{code}
```
\end{promptlisting}

\clearpage
\paragraph{Stage 5a --- Visual Judge.}
A vision model scores the rendered PNG and returns a strict JSON verdict;
\texttt{\{threshold\}} is the passing score.

\begin{promptlisting}
You are an expert reviewer of scientific-education diagrams. Your **primary duty** is to enforce **subject-matter correctness** strictly — any factual / knowledge error must trigger an outright rejection, no matter how visually polished the figure is.

Evaluate the following rendered scientific diagram.

## A. Subject-matter correctness (veto dimension)

**Compare every element in the figure against real-world / textbook standards.** If ANY of the following is true, reject:

1. **Shape / structure error** — an object's basic shape, composition, or geometric essence is wrong (e.g., the canonical morphology of an instrument / device / molecule / organ / celestial body / landform does not match reality, or two distinct things are drawn as the same)
2. **Topology / connectivity error** — adjacency, connection, input/output, or hierarchical relationships among parts are inverted or wrong
3. **Direction / order error** — vector direction, flow direction, temporal ordering, layer order, or cause-and-effect contradicts the facts
4. **Proportion / range error** — a key proportion that affects semantic classification is wrong (not artistic exaggeration, but a change of category)
5. **Mislabeling** — a label is attached to the wrong element, or a name impossible in the discipline is used
6. **Fabricated or missing key elements** — a part not belonging to the topic is added, or a part necessary for the topic is missing

Rule of thumb: review this as a textbook illustration — if a student studied from this figure, would they be misled? If yes → knowledge error.

The drawing description is reference only and **does not override real subject knowledge** — even if the description doesn't stress a detail, the figure must still be flagged when wrong.

## B. Visual clarity (0-10)

Layout, readability, non-crossing leaders, sharp text.

**Hard constraint — text legibility (violation forces clarity = 0, treated as veto)**:
- No two text spans may **overlap each other** (labels / formulas / axis numbers stacked on each other)
- Text and structural elements may **not overlap** (a label sitting on top of a \draw line / arrow / filled region, occluding the text or muddying the background)
- Text must **not be clipped** by frame / node boundaries; nothing extends off-canvas
If even one such overlap / occlusion / clipping is present, `clarity` **must be set to 0** and `overlap_issues` must record the location ("label X overlaps curve Y" / "label 'Na+' on top of threshold dashed line").

## C. Completeness (0-10)

Whether every element required by the description is present.

## Decision rules (must be obeyed strictly)

- If `knowledge_errors` is non-empty → `correctness` must be ≤ 4, and `pass` **must be false** (even if clarity and completeness are both 10)
- If `overlap_issues` is non-empty → `clarity` must be **= 0**, and `pass` **must be false**
- Otherwise `pass = (overall >= {threshold})`
- `feedback` must enumerate every knowledge error and every overlap issue with a **specific location** and a **direction for fixing**, so the regen prompt can act on it

Output strictly the following JSON (and nothing else):

{
  "knowledge_errors": ["<first error: element / location / why wrong>", "<second>", ...],
  "overlap_issues": ["<overlap 1: specific location, e.g., 'label A on top of curve B'>", ...],
  "correctness": <0-10; ≤ 4 when knowledge_errors is non-empty>,
  "clarity": <0-10; = 0 when overlap_issues is non-empty>,
  "completeness": <0-10>,
  "overall": <0-10>,
  "pass": <true/false>,
  "feedback": "<if pass=false: first restate every knowledge_errors and overlap_issues fix direction, then any other issues>"
}

Drawing description (reference only; cannot override subject facts):
{description}
\end{promptlisting}

\paragraph{Stage 5b --- Visual Regeneration.}
When the judge rejects, the model is shown the previous render plus the judge
feedback and asked to localize and fix. (Up to a configurable number of rounds.)

\begin{promptlisting}
You are a professional LaTeX/TikZ scientific-diagram developer.

The previous TikZ code compiled cleanly but the rendered result was rejected by quality review.
**The attached image is the actual render of the previous code** — use it to localize the issues and fix them.

Strict requirements:
1. Must be a complete standalone document
2. Target compiler is pdflatex
3. Output only the code (inside a code block)
4. Inspect the attached image carefully: which \draw / \node / coordinates correspond to the issues mentioned in the feedback (shape / position / overlap / labeling)? Localize before editing — do not rewrite the whole file.
5. Avoid repeating the previous version's mistake — if feedback says the shape was wrong, do NOT just nudge numbers; switch to a different drawing approach

{subject_hints}

Original drawing description:
{description}

Previous code:
```latex
{prev_code}
```

Quality feedback (must be addressed):
{feedback}

Output the improved complete TikZ code.
\end{promptlisting}

\paragraph{Per-Discipline Package Hints (\texttt{subject\_hints}).}
The same \texttt{\{subject\_hints\}} block is injected into Stages 1, 2, 4 and 5b.
It gives the model a curated list of available TikZ packages and drawing
conventions, which sharply reduces failures where the model invents a
non-existent package or hand-draws something a library already provides. One
block is shown per discipline.

\begin{promptlisting}
[Physics]
Available specialized packages (use as needed, not required):
- \usepackage{circuitikz} — circuit diagrams (resistor/capacitor/battery/ammeter/transistor/switch); use European symbol style
- \usetikzlibrary{optics} — optics (convex/concave lens, plane mirror, spherical mirror; combine with hand-drawn rays)
- \usepackage{pgfplots} — function plots, data plots, phase diagrams
- \usepackage{tikz-3dplot} — 3D coordinates, force decomposition, inclined planes
- \usepackage{tikz-feynman} — Feynman diagrams (high-energy physics only)
Native TikZ is fine for: free-body diagrams, motion trajectories, waveforms, magnetic/electric field lines.
Notes:
- Force decomposition must be along orthogonal axes; the resultant and components must be visually distinct
- Use arrows.meta arrowheads for ray direction in optics diagrams

[Biology]
Available specialized packages (use as needed, not required):
- \usepackage{chemfig} — molecular structures (sugars, amino acids, fatty acids, nucleotide backbones)
- \usepackage{forest} — phylogenetic trees, taxonomic trees
Native TikZ is fine for: cell structures, organ cross-sections, synapses, pedigrees, food webs.
Drawing conventions:
- Draw the outermost contour first (cell membrane / organ boundary), then the internal structures
- Distinguish each organelle/structure with its own fill color
- Labels in English; leader lines must not cross; place labels outside the element with thin lines pointing in
- Do NOT cram every structure into one coordinate — distribute them sensibly inside the outline

[Geography]
Available specialized packages (use as needed, not required):
- \usepackage{pgfplots} — contour lines, climate data, coordinate grids
- \usepackage{tkz-euclide} — geometry / projection
Native TikZ is fine for: geological cross-sections, plate tectonics, atmospheric circulation, water cycle, ocean currents.
Drawing conventions:
- Distinguish rock layers with patterns (\usetikzlibrary{patterns}; e.g., pattern=dots / north east lines) or semi-transparent fills
- Stratigraphy: in sedimentary rock, age decreases upward; mark fault dip with dashed lines / arrows showing hanging-wall vs footwall motion
- Draw circulation / currents as arrowed arcs; warm vs cold currents in warm/cool colors

[Chemistry]
Available specialized packages (use as needed, not required):
- \usepackage{chemfig} — organic / inorganic molecular structures (\chemfig{...}); handles bond angles, stereochemistry wedges, ring structures
- \usepackage[version=4]{mhchem} — inline formulas and equations \ce{H2SO4}, \ce{2H2 + O2 -> 2H2O}; auto subscripts/arrows
- \usepackage{pgfplots} — reaction curves, titration curves, energy profile diagrams, phase diagrams
Native TikZ is fine for: lab apparatus (flask / condenser / separating funnel / alcohol burner), reaction-mechanism flows.
Drawing conventions:
- Always use chemfig for molecular structures; do NOT assemble molecules from bare \node
- Always use \ce{} for chemical equations; do NOT hand-write subscripts
- Apparatus diagrams must show clear interfaces (rubber stopper / tube direction); label heat source and condenser water in/out clearly
- Reaction curves must label axis meaning (time / volume / pH etc.) and key points (equivalence point, inflection)

[CS (Computer Science)]
Available specialized packages (use as needed, not required):
- \usetikzlibrary{arrows.meta,positioning,shapes,calc,chains} — node placement, arrow styles
- \usepackage{forest} — binary trees, expression trees, directory trees
- \usepackage{tikz-qtree} / qtree — syntax trees, parse trees
Native TikZ is fine for: linked lists / stacks / queues / hash tables, graphs (directed / undirected), state machines, sequence diagrams, flowcharts, array / memory layouts.
Drawing conventions:
- Use rectangle split (\usetikzlibrary{shapes.multipart}) to split a node into data / next fields
- Pointer arrows use -{Stealth}; use a slash fill or ∅ for NULL/None pointers
- Keep adjacent node spacing consistent — use positioning's `right=of ...` rather than hard-coded coordinates
- Graphs: directed = arrows, undirected = plain lines; weights labeled at edge midpoints

[Economics]
Available specialized packages (use as needed, not required):
- \usepackage{pgfplots} — supply/demand curves, IS-LM, AD-AS, Phillips curve, utility curves (axis lines=middle puts the origin at the center)
- \usetikzlibrary{arrows.meta,intersections,calc} — curve intersections, shifted curves
Native TikZ is fine for: game trees, production possibilities frontier, cost-benefit geometry.
Drawing conventions:
- Axes must label variable names (P, Q, r, Y, ...) and units
- Project equilibrium points to the axes with dashed lines, label P*, Q* etc. on the axes
- When shifting a curve, draw both old and new curves and distinguish with D, D' / S, S'; show shift direction with a small arrow
- Use semi-transparent fills for shaded regions (consumer surplus / producer surplus / deadweight loss)
\end{promptlisting}

\subsection{T2I Data Filtering Pipeline}

We collect open-source datasets, including K12~\cite{waltonfuture_k12}, MV-MATH~\cite{wang2025mv}, MAVIS~\cite{zhang2024mavismathematicalvisualinstruction}, and ChemEval~\cite{huang2026chemeval}, comprising \textbf{2.5M} samples in total, to construct an automated filtering pipeline for text-to-image data. First, we perform global pHash~\cite{zauner2010phash} deduplication on the original images and remove images that are duplicate or highly similar to those in the target benchmarks (GenExam~\cite{wang2025genexam}, GRADE~\cite{liu2026grade}, and RISE~\cite{zhao2025envisioning}), to ensure the independence of the evaluation sets. We then discard images with small sizes (short size < 512 pixels) to ensure that the retained samples contain sufficient visual information.

After the basic cleaning stage, we further apply an automated discipline-image filtering pipeline designed to retain visually informative, discipline-relevant educational diagrams while excluding text-dominant or natural-image samples. The pipeline operates in three stages. 

\textbf{OCR Pre-Screening.} In the first stage, we perform lightweight OCR-based pre-screening before invoking the multimodal model. For each image, the OCR~\cite{paddleocr3} module estimates the total number of detected characters, the number of text boxes, and the text-area ratio, i.e., the proportion of image area occupied by text. Images are directly rejected at this stage if the prompt contains Chinese characters, if the detected text count exceeds \textbf{280} characters, or if text occupies more than \textbf{13\%} of the image area, since such samples are likely to be language-dependent or overly text-centric rather than suitable for visual generation.

\textbf{Multimodal Quality Assessment.} For the remaining images, we use Qwen3-VL-8B-Instruct~\cite{Qwen3-VL} to conduct structured multi-dimensional quality assessment. The model outputs judgments for whether the image is a real photograph (\textbf{is\_real\_photo: 0/1}), its suitability as a discipline-specific educational image (\textbf{discipline\_image\_score: 0--10}), its dependence on textual content (\textbf{text\_richness\_score: 0--10}), whether it belongs to an abstract/synthetic diagram domain rather than a natural-image domain (\textbf{image\_domain\_score: 0/1}), its visual complexity (\textbf{image\_complexity\_score: 0--10}), and the amount of domain knowledge required to understand it (\textbf{discipline\_knowledge\_score: 0--10}). In addition, the model predicts a knowledge-difficulty level, a coarse image type, a fine-grained image subtype, a discipline label, and a free-text rationale. The coarse image types are defined over a closed set including diagram, chart\_plot, geometry, formula, map, molecule, circuit, screenshot, photo, mixed, and other, while the fine-grained taxonomy further distinguishes more than \textbf{90} educational subcategories, such as function graphs, geometric transformations, force analysis diagrams, molecular structures, optical paths, algorithm flowcharts, and geological maps. Table~\ref{tab:score_distribution} summarizes the score distributions of the four main evaluation dimensions used in downstream filtering, namely discipline\_image\_score, text\_richness\_score, image\_complexity\_score, and discipline\_knowledge\_score.

\begin{table*}[htbp]
  \caption{Score distribution across four main evaluation dimensions. (Total $n$ = 582,789)}
  \centering
  \setlength{\tabcolsep}{8pt}
  \renewcommand{\arraystretch}{1.3}
  \resizebox{0.7\linewidth}{!}{
  \begin{tabular}{ccccc}
  \toprule
  \textbf{Score} & \textbf{Discipline Image} & \textbf{Text Richness} & \textbf{Image Complexity} & \textbf{Discipline Knowledge} \\
  \midrule
  0  & 436 (0.07\%)     & 23,677 (4.06\%)  & -                & -                 \\
  1  & 145 (0.02\%)     & 2 (0.00\%)       & 4,846 (0.83\%)   & 2,608 (0.45\%)    \\
  2  & 145 (0.02\%)     & 2 (0.00\%)       & 1,017 (0.17\%)   & 365 (0.06\%)      \\
  3  & 383 (0.07\%)     & 23 (0.00\%)      & 94,524 (16.22\%) & 2,428 (0.42\%)    \\
  4  & 83 (0.01\%)      & -                & 73,721 (12.65\%) & 1,020 (0.18\%)    \\
  5  & 20 (0.00\%)      & -                & 31,878 (5.47\%)  & 5,577 (0.96\%)    \\
  6  & 487 (0.08\%)     & -                & 332,643 (57.08\%)& 23,622 (4.05\%)   \\
  7  & 46 (0.01\%)      & 26,657 (4.57\%)  & 640 (0.11\%)     & 10,353 (1.78\%)   \\
  8  & 3,521 (0.60\%)   & 54 (0.01\%)      & 43,475 (7.46\%)  & 320,967 (55.07\%) \\
  9  & 59,552 (10.22\%) & 17 (0.00\%)      & 1 (0.00\%)       & 51,839 (8.90\%)   \\
  10 & 518,116 (88.90\%)& 532,357 (91.35\%)& 44 (0.01\%)      & 164,010 (28.14\%) \\
  \bottomrule
  \end{tabular}
  }
  \label{tab:score_distribution}
\end{table*}

\textbf{Rule-Based Filtering.} We then apply rule-based post-filtering that combines OCR statistics with model predictions. An image is rejected if it is classified as a real photograph, if its discipline-image score is \textbf{below 4}, if its text-richness score is \textbf{below 10}, if its image-domain score is \textbf{not exactly 1}, if its complexity score is \textbf{below 4}, or if its discipline-knowledge score is \textbf{below 4}. We further restrict the accepted set to a whitelist of educational visual formats, namely diagram, chart\_plot, geometry, formula, map, molecule, and circuit, while excluding screenshots, photographs, mixed-format images, and uncategorized cases. This design intentionally imposes exact constraints on text richness and image domain, so that the final dataset consistently contains text-free, non-photographic, structurally clear, visually reproducible, and academically meaningful discipline images.

\textbf{Dataset Construction.} After filtering, the retained samples are further grouped by discipline categories such as mathematics, physics, chemistry, biology, geography, computer science, history, music, and sports for downstream discipline-specific text-to-image generation.

\subsubsection{Prompt of Multimodal Quality Assessment}
\label{appendix:multimodal-quality-assessment-prompt}
This section reproduces the prompt used for Multimodal Quality Assessment in the T2I Data Filtering stage. The model uses the image as the primary signal and the paired instruction text as supporting context, then returns a strict JSON record.

\begin{promptlisting}
Score this image for dataset filtering. Use the image first, and use the paired instruction as supporting context.

instruction:
```
You are scoring educational images for dataset filtering.

You must output strict JSON only.

Input:
- An image
- Its paired instruction text, when available

Your job:
1. Decide whether the image is a real photograph.
2. Score whether it is a useful subject-learning image.
3. Score text richness, image domain, image complexity, and subject knowledge need.
4. Classify the image type from a closed set.
5. Assign one subject if the image is clearly academic.

Closed-set subjects:
Mathematics, Physics, Computer Science, Chemistry, Geography, Biology, Music, History, Sports, Economics

Closed-set difficulty labels:
Common Knowledge, Middle School, High School, University, Professional/Specialized

Closed-set image types:
diagram, chart_plot, geometry, formula, table, map, molecule, circuit, screenshot, photo, mixed, other

Closed-set fine-grained image types:
- Mathematics: Plane Geometry Fill-in Values, mathcanvas, Plane Image Operations by Vectors, Triangle Median/
  Altitude/Midsegment, Point-Line Perpendicularity/Parallelism, Solid Geometry, Function Zeros/Poles/Derivatives/
  Tangents/Trigonometric Functions/Feasible Region, Statistics, Generate Graph from Function, Feasible Region,
  Function Periodicity
- Chemistry: Chemical Equation Fill-in, Molecular Formula (Bond-line Notation), Atomic Structure, Experimental
  Apparatus/Reaction Curves, Ball-and-Stick Model/Van der Waals Surface Model, Intermolecular Reaction Mechanism
  Diagram, Diverse Molecular Representations, Reaction Equation
- Biology: Flowchart, Molecule/Reaction Formula/Reaction Curve, Life Cycle, Food Web, DNA Structure and Processes,
  Cell/Organ/Tissue Structure, Evolution, Genetic Map/Pedigree, Molecular Structure, Neuron Action Potential Curve
  /Pharmacological Curve/Stress Response Diagram, Reaction Diagram
- Computer Science: Linked List, Neural Network/Convolution Operation, Tree/State Machine/Graph, Machine Learning
  Courseware Illustrations, Computer System, Graph Data Structures: Heap/Binary Tree/Graph/Markov Chain, Neural
  Network, Digital Circuit, Pseudocode Flowchart
- Geography: Map (Ocean Current/Wind/Plate Tectonic Direction) with Arrows, Structure Diagram Fill-in, Astronomy,
  Earth's 3D Structure, Geological Structure
- History: Map Administrative Region Fill-in, Route Map, Timeline, Transition Curve
- Music: Single Note, Staff Notation, Rolling Staff, Key Signature/Clef/Note Rendering for Simple Content, Single
  Note on Staff
- Physics: Three-View Drawing of Parts, Light Path Diagram, Force Analysis, Circuit Diagram, Thermodynamics/
  Mechanics Curve, Experimental Apparatus, Analog Circuit, Light Path, Physical Curves
- Sports: Board Games, Muscle, Strategy and Tactics, Food Pie Chart/Real Object Photo/Blood Sugar Changes
- Economics: Economic Curves, Economic Statistical Charts

Scoring definitions:
- subject_image_score: 0-10
  0 = not useful for subject-image generation, 10 = very suitable
- text_richness_score: 0-10
  0 = pure text/table-only style image with no meaningful non-text visual content
  10 = no text at all
  intermediate values depend on how much the image meaning relies on text
- image_domain_score: 0-1
  0 = natural image / pathology image / body scan image / MRI / CT / X-ray / real capture
  1 = non-natural schematic / chart / diagram / symbolic image
- image_complexity_score: 1-10
  1 = extremely complex and hard to reproduce
  10 = very simple
- subject_knowledge_score: 1-10
  1 = no subject knowledge needed
  10 = strongly requires subject knowledge and reasoning
- image_type_score: 1-5
  confidence that the chosen image_type is correct

Important distinctions:
- Mark is_real_photo = true only for real-world photos / scans / realistic captures.
- If the image is not clearly academic, subject can be null.

Return exactly this JSON schema:
{
  "is_real_photo": false,
  "subject_image_score": 8,
  "text_richness_score": 7,
  "image_domain_score": 1,
  "image_complexity_score": 6,
  "subject_knowledge_score": 8,
  "image_type": "diagram",
  "image_type_score": 5,
  "fine_grained_image_type": "Schematic Diagram",
  "subject": "Physics",
  "knowledge_difficulty_score": 4,
  "knowledge_difficulty_label": "University",
  "reason": "short reason"
}

Rules:
- subject_image_score and text_richness_score must be integers in [0, 10].
- image_domain_score must be 0 or 1.
- image_complexity_score and subject_knowledge_score must be integers in [1, 10].
- knowledge_difficulty_score and image_type_score must be integers in [1, 5].
- image_type must be one of the types in 'Closed-set fine-grained image types'.
- subject must be null or one of: Mathematics, Physics, Computer Science, Chemistry, Geography, Biology, Music, History, Physical Education
- knowledge_difficulty_label must be one of: Common Knowledge, Middle School, High School, University, Professional/Specialized
- Do not output markdown.
- Do not output extra keys.
```

\end{promptlisting}

\clearpage
\section{Training Details}
\label{sec:training_details}

The reported models use the same training code path in DiffSynth-Studio. Qwen3-VL-8B is used as the reasoning module and is kept frozen. We only optimize LoRA adapters attached to the DiT backbone of the image generator; the original generator weights, text encoder, and VAE remain frozen. All image generators are loaded in bfloat16, and Accelerate is launched with 8 processes on one 8-GPU node. Since the dataloader uses one sample per process and gradient accumulation is 1, the effective batch size is 8 for both T2I generation and editing.

\paragraph{Text-to-Image Generation.}
The T2I model starts from Qwen-Image-2512 and is trained on the final T2I split of \datasetname.
We train for one effective pass over this split, corresponding to 43,993 optimizer steps with the effective batch size above. The LoRA rank is 128. We use AdamW with learning rate $5\times10^{-6}$, weight decay 0, and a constant learning-rate schedule without warmup. Checkpoints are saved every 1,000 optimizer steps and only the latest checkpoint is kept during training. To reduce the tendency to overfit plain white diagram backgrounds, we apply a background loss multiplier of 0.2 to pixels whose RGB distance from the border-estimated background color is at most 24; foreground pixels keep unit weight. The border color is estimated from a 6\% border band.

\paragraph{Image Editing.}
The editing model starts from Qwen-Image-Edit-2511 and is trained on the final editing split of \datasetname after validity filtering and duplicate removal. 
We train for one effective pass, corresponding to 23,384 optimizer steps. The LoRA rank is 64. We use AdamW with learning rate $2\times10^{-6}$, weight decay 0, and a constant learning-rate schedule without warmup. The editing dataloader uses \texttt{image,edit\_image} as the paired image fields and passes \texttt{edit\_image} as an extra model input. We enable the Qwen-Image-Edit \texttt{zero\_cond\_t} option. Checkpointing follows the same setting as T2I: every 1,000 optimizer steps with a single retained checkpoint.

\paragraph{Resolution and Data Processing.}
For both tasks, we leave the fixed height and width unset and use dynamic-resolution training. Each image is resized/cropped by the dataset operator with a maximum pixel budget of 1,572,864 pixels, and the processed height and width are aligned to multiples of 16. The dataset repeat factor is 1. The training loss is the standard FlowMatch SFT objective used by the Qwen-Image training script; timestep sampling uses 1,000 training timesteps from the pipeline scheduler.

\paragraph{LoRA Placement.}
For both models, LoRA adapters are inserted into the DiT attention projections, including the query, key, value, added query/key/value, and output projections. For T2I generation, we additionally adapt the image/text MLP output projections and modulation layers. This gives the T2I model more capacity for domain-specific layout, notation, and text-heavy diagram structure, while the editing model uses the more conservative attention-only adapter set to preserve the source image.

\end{document}